\documentclass{article}
\usepackage{ijcai25}

\usepackage{times}
\usepackage{soul}
\usepackage{url}
\usepackage[dvipsnames]{xcolor}
\usepackage[hidelinks, colorlinks=true, linkcolor=RoyalBlue, citecolor=RoyalBlue]{hyperref}
\usepackage[utf8]{inputenc}
\usepackage[small]{caption}
\usepackage{graphicx}
\usepackage{amsmath}
\usepackage{amsthm}
\usepackage{booktabs}
\usepackage{algorithm}
\usepackage{algorithmic}
\usepackage[switch]{lineno}
\usepackage{witharrows}

\usepackage{stackengine}
\newcommand{\mymathbox}[4]{%
  \stackengine{2pt}{\colorbox{#1!20}{$\displaystyle#3$}}{\scriptsize\textcolor{#1!80!black}{#4}}{O}{c}{F}{T}{S}%
}
\definecolor{grayLL}{RGB}{250, 250, 250}
\definecolor{grayL}{RGB}{227, 227, 227}

\definecolor{graysampling}{RGB}{179,179,179}
\definecolor{greenarrow}{RGB}{137,188,0}

\definecolor{intBlueL}{RGB}{117,121,182}
\definecolor{intBlueLL}{RGB}{223,224,236}
\definecolor{intGrayL}{RGB}{135,135,135}
\definecolor{intGrayLL}{RGB}{227,227,227}
\definecolor{intCrimL}{RGB}{112, 59, 84}
\definecolor{intCrimLL}{RGB}{222,212,217}
\definecolor{intTealL}{RGB}{89,143,131}
\definecolor{intTealLL}{RGB}{218,229,226}

\usepackage{colortbl}
\usepackage[misc]{ifsym}
\definecolor{rred}{RGB}{245, 152, 153}
\definecolor{oorange}{RGB}{253, 205, 154}
\definecolor{yyellow}{RGB}{248,244,140}
\definecolor{ydyellow}{RGB}{225,215,9}
\definecolor{ggrean}{RGB}{50,205,50}
\usepackage{amsmath}
\usepackage{amssymb}
\usepackage{booktabs}

\title{Optimized View and Geometry Distillation from Multi-view Diffuser}

\author{
Youjia Zhang$^1$
\and
Zikai Song$^1$\and
Junqing Yu$^1$\and
Yawei Luo$^2$\And
Wei Yang$^{1 \dag}$\\
\affiliations
$^1$Huazhong University of Science and Technology\\
$^2$Zhejiang University\\
\emails
\{youjiazhang, weiyangcs\}@hust.edu.cn
}

\begin{document}
\maketitle
\begin{abstract}
Generating multi-view images from a single input view using image-conditioned diffusion models is a recent advancement and has shown considerable potential. However, issues such as the lack of consistency in synthesized views and over-smoothing in extracted geometry persist. Previous methods integrate multi-view consistency modules or impose additional supervisory to enhance view consistency while compromising on the flexibility of camera positioning and limiting the versatility of view synthesis. In this study, we consider the radiance field optimized during geometry extraction as a more rigid consistency prior, compared to volume and ray aggregation used in previous works. We further identify and rectify a critical bias in the traditional radiance field optimization process through score distillation from a multi-view diffuser. We introduce an \textbf{Unbiased Score Distillation (USD) } that utilizes unconditioned noises from a 2D diffusion model, greatly refining the radiance field fidelity. We leverage the rendered views from the optimized radiance field as the basis and develop a two-step specialization process of a 2D diffusion model, which is adept at conducting object-specific denoising and generating high-quality multi-view images. Finally, we recover faithful geometry and texture directly from the refined multi-view images. Empirical evaluations demonstrate that our optimized geometry and view distillation technique generates comparable results to the state-of-the-art models trained on extensive datasets, all while maintaining freedom in camera positioning. Source code of our work is publicly available at: \url{https://youjiazhang.github.io/USD/}.
\end{abstract}

\let\thefootnote\relax\footnotetext{$^\dag$Corresponding author.}

\section{Introduction}
\label{sec:intro}
Traditionally, the process of generating a three-dimensional model from a singular image necessitates extensive and meticulous efforts by highly skilled artists.
However, recent advancements in neural networks, particularly through the adaptation of 2D diffusion models for 3D synthesis, have rendered the conversion of a single image into a 3D object feasible. 
The early breakthrough comes from the text to 3D domain, where DreamFusion~\cite{poole2022dreamfusion} and Score Jacobian Chaining (SJC)~\cite{wang2023score} proposes a Score Distilling Sampling (SDS) strategy to distill the scores learned by 2D diffusion models from large-scale images to optimize a Neural Radiance Field (NeRF)~\cite{mildenhall2020nerf}, circumventing the need for 3D data. %During SDS process, a random-scaled noise is injected into NeRF render images, and then denoised by a 2D diffusion model to provide supervision.
Successive approaches further improve the quality and diversity of generated geometries from textural prompts~\cite{wang2023prolificdreamer,lin2023magic3d,chen2023fantasia3d}. Particularly, RealFusion~\cite{melas2023realfusion} migrates the scheme to generate plausible 3D reconstruction matches to a single input image via textual inversion adapted supervision.

%Many studies follow this path, Magic3D~\cite{prolific} adopts a two-stage optimization framework to improve the quality of the generated 3D mesh. ProlificDreamer~\cite{prolific} proposes the Variational Score Distillation (VSD) scheme to overcome the oversmoothness issue.

More relevantly, 3DiM~\cite{watson2022novel} and MVDream~\cite{MVDream} develop a pose-conditional image-to-image diffusion model, which generates the novel view at a target pose from a source view. Zero-1-to-3~\cite{liu2023zero} adopts a similar framework and learns control of viewpoints through a synthetic dataset and demonstrates zero-shot generalization to in-the-wild images. Though Zero-1-to-3 demonstrates plausible novel views, they are not multi-view consistent and the geometry distilled from SDS tends to be oversmoothed. To enhance the multi-view consistency, SyncDreamer~\cite{liu2023syncdreamer} devises a volume-encoded multi-view noise predictor to share information across different views. Wonder3D~\cite{long2023wonder3d} predicts the multi-view color images along with their normal maps from a cross-domain diffusion model. Though enhancing the multi-view consistency of image generation, these methods compromise the flexibility of camera positioning and only allow synthesis for a limited number of views.
%these methods can recover 3D shapes from the generated multi-view images.

In this study, we observe that the predicted unconditional noise from the multi-view diffuser, \textit{i.e.}, the Zero-1-to-3 model, appears to be biased. That is, even if we only add very low-level noise to a normal image and use the unconditional noise predicted by a Zero-1-to-3 model for denoising, the result still tends to deviate greatly from the original image. We analyze and rectify the critical bias by using an unconditioned noise from a pre-trained 2D diffusion model and greatly refining the geometry fidelity.
%in the geometry extraction using the Zero-1-to-3 model~\cite{xxx} and SDS. We introduce an unbiased xxx that utilizes unconditioned noises from a 2D diffusion model, greatly refining the geometry fidelity. 
%Moreover, images rendered from geometry tend to be blurry. 
Moreover, previous approaches use either 3D volume~\cite{liu2023syncdreamer} or ray aggregation~\cite{tseng2023consistent} to share information across views. We consider the radiance field as the consistency prior, and encourage the generated multi-view images to be consistent with the NeRF renderings. 
We develop a two-step specialization process of a 2D diffusion model, which is adept at conducting target-specific denoising and generating high-quality multi-view images from NeRF renderings. We then further use the refined views to generate the geometry and texture from NeuS~\cite{wang2021neus}, and in the meanwhile enforce input view consistency using view score distillation. Our approach generates comparable-quality of multi-view images and geometry to the SOTA approaches, including SyncDreamer and Wonder3D, without enforcing any restriction on camera poses. Consequently, we posit that our approach offers superior adaptability and effectiveness in addressing the challenges associated with generating consistent and high-quality multi-view imagery and geometry, affirming its substantial potential for widespread application in relevant fields.

\begin{figure}[t]
    \centering
    \includegraphics[width=1.0\linewidth]{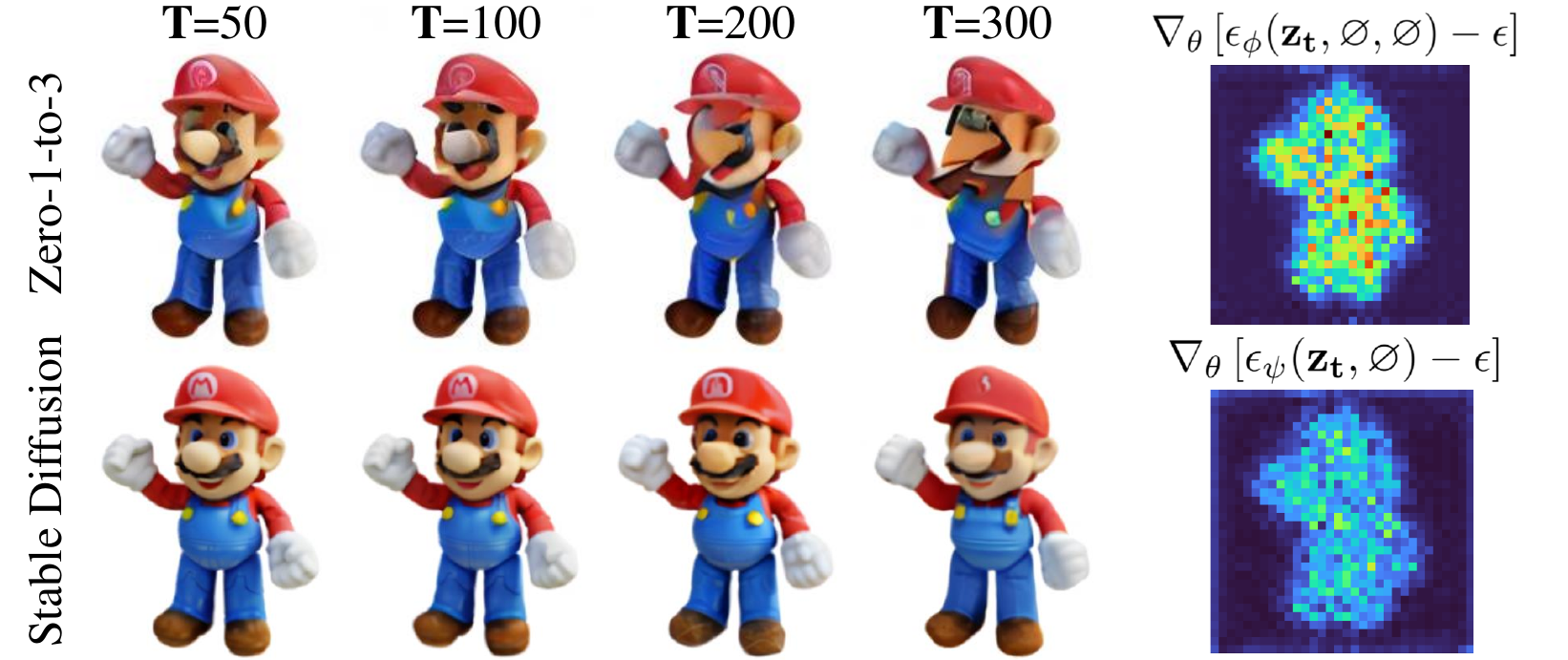}
    \caption{The unconditional noise predicted by Zero-1-to-3 model tends to be biased. As a demonstration, we use the `$\textit{Mario}$' image as a toy example and add various levels of noise to the image (larger $\mathbf{T}$ means more noise has been added). We use the predicted unconditional noise to recover the original image from noisy input and find the results of Zero-1-to-3 deviate from the input image greatly even for very small amount of noise. The right sub-figure shows the averaged difference between the predicted noise and the added noise.}
    \label{uncond_noise}
\end{figure}

\begin{figure*}[t]
    \centering
    \includegraphics[width=1.0\linewidth]{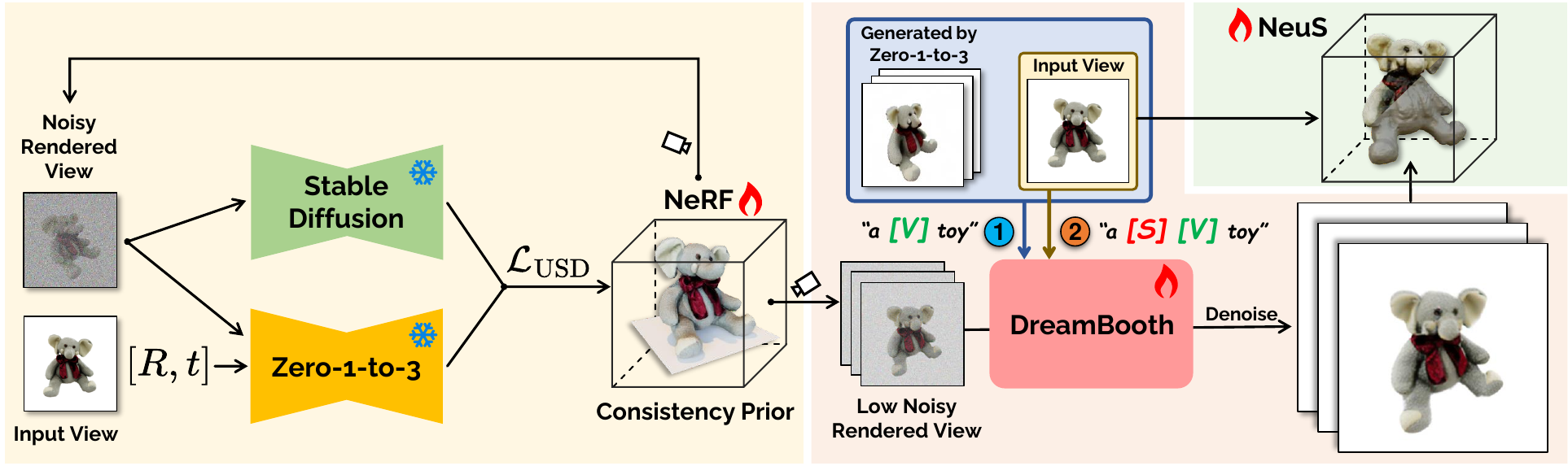}
    \caption{
     The overall pipeline of our approach. We first use our Unbiased Score Distillation to extract an optimized underlying radiance field. And then we use the NeRF as our consistency prior, \textit{i.e.}, the generated views should be consistent with the NeRF renderings. We propose a two-stage specialization scheme to obtain a specified DreamBooth specifically for the target. We then denoise the NeRF renderings to obtain high-quality views and subsequently use NeuS technique to recover the geometry. Our optimized scheme generates comparable, sometimes better particularly for irregular camera poses, results to the SOTA works without training on large-scale data.
    }
    \label{pipeline}
    \vspace{-8pt}
\end{figure*}

\section{Related Work}

\noindent \textbf{2D Diffusion Models for 3D Generation.}
% ho2020denoising
Recent advancements in 2D diffusion models \cite{ho2020denoising,rombach2022high,croitoru2023diffusion} and large-scale visual language models, notably the CLIP model \cite{radford2021learning,song2024autogenic}, have catalyzed new approaches for generating 3D assets. Pioneering efforts such as DreamFusion \cite{poole2022dreamfusion} and SJC \cite{wang2023score} have developed methods for transforming 2D text into images, subsequently facilitating the generation of 3D shapes from text. This approach has inspired a range of subsequent studies that adopt a shape-by-shape optimization scheme. Additionally, the integration of 2D diffusion models with robust vision language models, especially CLIP \cite{radford2021learning}, has emerged as a significant exploration in the generation of 3D assets \cite{xu2023dream3d}.
The typical methodology involves optimizing a 3D representation, such as NeRF, mesh, or SDF, and then utilizing neural rendering to generate 2D images from various viewpoints. These images are processed through 2D diffusion models or the CLIP model to calculate SDS losses, which guide the optimization of the 3D shape. Building on the foundations laid by DreamFusion and SJC, numerous works have enhanced text-to-3D distillation methods in various aspects. Notably, Magic3D \cite{lin2023magic3d} develops a two-stage coarse-to-fine optimization framework for high-resolution 3D content generation, and ProlificDreamer \cite{wang2023prolificdreamer} proposes a Variational Score Distillation (VSD) for generating highly detailed geometry.
However, challenges such as low efficiency and the multi-face Janus problem, where optimized geometry tends to produce multiple faces due to the lack of explicit 3D supervision, remain prevalent. Furthermore, some works~\cite{radford2021learning} have applied this distillation pipeline in single-view reconstruction tasks. While these methods have achieved impressive results, they often require extensive time for textual inversion and NeRF optimization, without always guaranteeing satisfactory outcomes. In contrast with the 2D diffusion to 3D extension, which is ignorant to multi-view consistency, our method focuses on the multi-view diffuser technique, which predicts the noises for novel views, and inherently avoids the Janus problem.

\noindent \textbf{Multi-view Diffusion Models.}
% tewari2023diffusion
In light of the complexities involved in ensuring the integrity of generated 3D content, recent efforts~\cite{watson2022novel,gu2023nerfdiff,deng2023nerdi,tseng2023consistent,chan2023generative,yu2023long,tang2023mvdiffusion,liu2023deceptive,tewari2023diffusion} have explored the feasibility of directly generating novel views from a single image input. Notably, the Scene Representation Transformer \cite{sajjadi2022scene} extends the vision transformer to image sets, enabling global information integration for 3D reasoning. Similarly, 3DiM \cite{watson2022novel} develops a pose-conditional image-to-image diffusion model, translating a single input view into consistent and sharp completions across multiple views. A seminal work in this area, Zero-1-to-3 \cite{liu2023zero}, utilizes a similar network structure, trained on a large-scale synthetic 3D dataset, demonstrating notable generalizability.
Recent advancements, such as One-2-3-45 \cite{DBLP:conf/cvpr/Xu0WFWW23}, have leveraged the generalizable neural reconstruction method SparseNeuS \cite{long2022sparseneus} to directly produce 3D geometry from images generated by Zero-1-to-3. While this approach is more efficient and alleviates the Janus (multi-head) problem, it tends to produce lower-quality results with reduced geometric detail.
In a different vein, SyncDreamer \cite{liu2023syncdreamer} concentrates on object reconstruction, generating images in a single reverse process and utilizing attention to synchronize states among views. This contrasts with Viewset Diffusion \cite{szymanowicz2023viewset}, which requires predicting a radiance field. SyncDreamer solely relies on attention for synchronization, fixing the viewpoints of generated views to enhance training convergence.
A significant trend in recent research has been the design of various functional attention layers. Consistent123 \cite{weng2023consistent123} employs a shared self-attention layer, where all views query the same key and value from the input view. ConsistNet \cite{yang2023consistnet} introduces two sub-modules: a view aggregation module and a ray aggregation module, to extract features consistent across multiple views. MVDream \cite{MVDream} utilizes 3D self-attention.
Wonder3D \cite{long2023wonder3d} goes a step further by not only outputting multi-view images but also outputting the normal map for each perspective, using cross-domain attention to maintain the consistency of the 3D structure. Both SyncDreamer and Wonder3D, resulting from fixed view outputs, exhibit sensitivity to the camera angle of the input image. In contrast, our method can reconstruct accurate 3D structures from inputs captured at varying camera angles. While existing approaches have focused on developing novel modules for enforcing multi-view consistency, our work demonstrates that an optimized distillation strategy can yield views and geometries comparable to models trained on large-scale datasets. This insight may inspire further exploration in improved strategies for geometry and view extraction.

\section{Method}

Our objective is to synthesize consistent multi-view images and high-quality geometric representations from a single input image. Notable prior work, such as the Zero-1-to-3, has demonstrated impressive results by utilizing an image and camera pose-conditioned diffusion model. However, this approach encounters limitations, particularly in terms of generating inconsistent multi-view images and a tendency for over-smoothing in the geometric output. The state-of-the-art research, including SyncDreamer~\cite{liu2023syncdreamer} and Wonder3D~\cite{long2023wonder3d}, addresses these challenges by incorporating additional modules for consistency or employing normal supervision. However, this often compromises the flexibility of positioning the target camera at will.
Contrastingly, our work adopts a distinct methodology, showcasing that comparable quality in views and geometry can be achieved through a meticulously crafted distillation strategy. Central to our approach is the insight that the unconditional noise predictions from Zero-1-to-3 are inherently biased. We propose the utilization of unconditional noise from the Stable Diffusion~\cite{takagi2023high} model to rectify this issue, as elaborated in Sec.~\ref{sec:Unbiased Multi-view Score Distillation}. Our method, termed Unbiased Score Distillation (USD), significantly enhances the quality of the radiance field relative to previously used SDS/SJC methods.
Furthermore, we employ the optimized NeRF~\cite{mildenhall2020nerf} as a consistency prior, in contrast to previous implicit constraints such as 3D volume or ray aggregation. We ensure that the generated views and geometry align coherently with the distilled NeRF to achieve consistency. We posit that specializing a diffusion model to denoise the target object is crucial. To this end, we implement the DreamBooth~\cite{ruiz2023dreambooth} technique and engage in a two-stage fine-tuning process, detailed in Sec.~\ref{sec:Consistent View and Geometry Distillation}. The first stage involves using multi-view images from Zero-1-to-3 as positive samples, contrasting them with text-prompt generated images as negative samples for image style learning. In the subsequent stage, the input image serves as the positive sample, with all Zero-1-to-3 generated images as negatives, focusing on learning finer details.
Subsequently, we introduce a low level of noise into the NeRF renderings and employ the fine-tuned diffusion model for denoising. The final step involves applying the NeuS technique for mesh reconstruction, while in the meanwhile using View Score Distillation to ensure input view consistency. Fig.~\ref{pipeline} illustrates the complete pipeline of our methodology.

\begin{figure*}[t]
    \centering
    \includegraphics[width=0.99\linewidth]{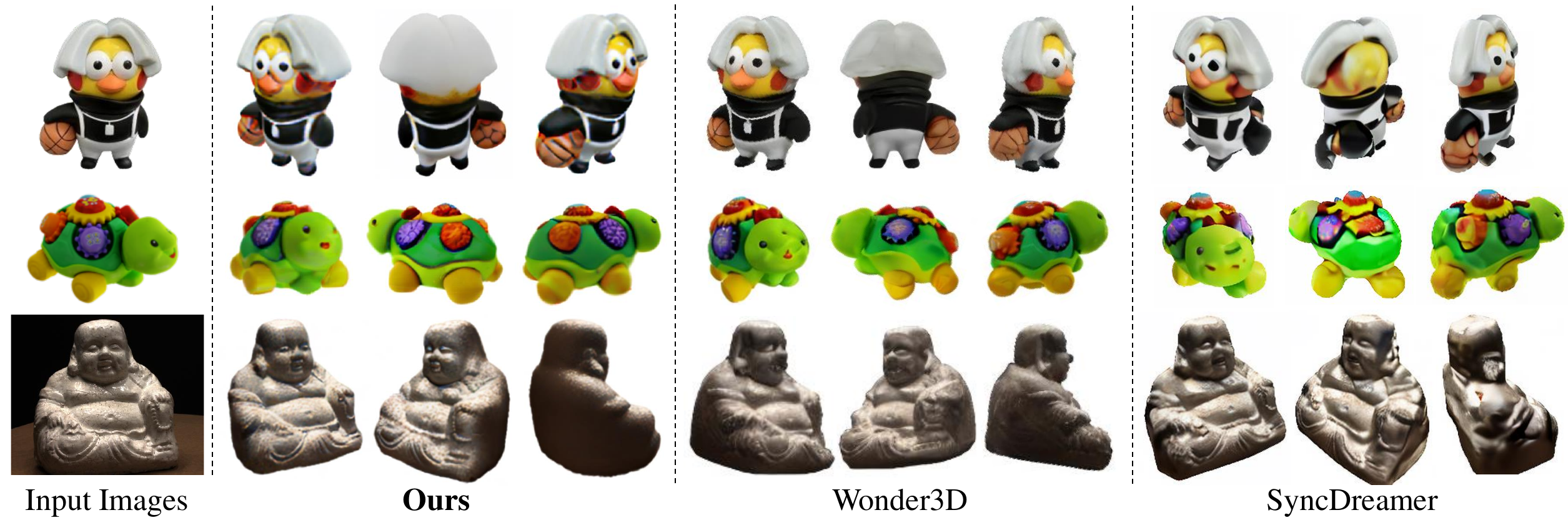}
    \caption{
    The qualitative comparisons with baseline models on multi-view color images. Our approach generates consistent multi-view images while preserving the image details. 
    }
    \label{color-compare}
    \vspace{-8pt}
\end{figure*}

\subsection{Unbiased Score Distillation}
\label{sec:Unbiased Multi-view Score Distillation}

In this section, we highlight the significant bias in Zero-1-to-3's unconditional noise due to insufficient unconditional training and object-level dataset bias, affecting geometry quality in SDS-based distillation, both theoretically and empirically. We then propose a rectification method.

\begin{figure*}[t]
    \centering
    \includegraphics[width=0.9 \linewidth]{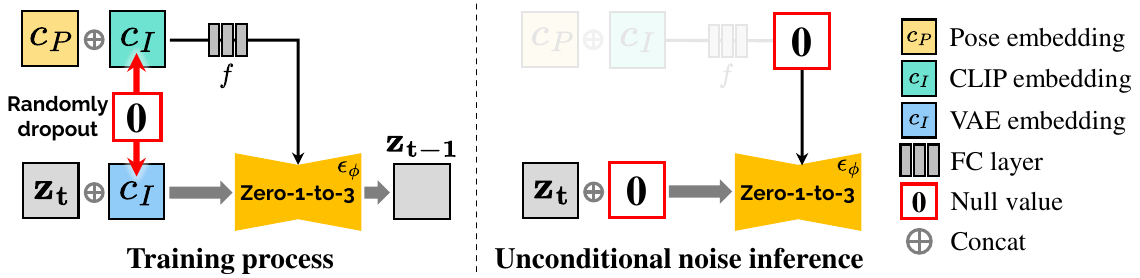}
    \caption{
    Training and inference process of Zero-1-to-3. Training for predicting unconditional noise involves setting the $c_I$ conditions to 0 at regular intervals.
    }
    \label{Zero123}
    % \vspace{-8pt}
\end{figure*}

\noindent \textbf{Bias in Unconditional Noise.}
As a multi-view diffuser, Zero-1-to-3 predicts noises of a target image latent $\mathbf{z}$ given two conditions: the input image $c_I$ and \textit{relative} camera pose $c_P$, \textit{i.e.}, it learns the probability distribution $P(\mathbf{z}|c_I,c_P)$ for image latent $\mathbf{z}$. We apply {\small \textit{Bayes' Rule}} to decompose the conditional probability:
\begin{equation}
P(\mathbf{z}|c_I,c_P)=\frac{P(\mathbf{z},c_I,c_P)}{P(c_I,c_P)}=\frac{P(c_I|c_P,\mathbf{z})P(c_P|\mathbf{z})P(\mathbf{z})}{P(c_I,c_P)}  
\end{equation}
\noindent Diffusion models estimate the score ~\cite{hyvarinen2005estimation} of the data distribution, \textit{i.e.}, the derivative of the log probability, giving us the following expression:
\begin{equation}
\begin{split}
\nabla_\mathbf{z}\mathrm{log}(P(\mathbf{z}|c_I,c_P)) = & \nabla_\mathbf{z}\mathrm{log}(P(c_I|c_P,\mathbf{z})) + \\
& \nabla_\mathbf{z}\mathrm{log}(P(c_P|\mathbf{z})) + \\
& \nabla_\mathbf{z}\mathrm{log}(P(\mathbf{z}))
\end{split}
\end{equation}
\noindent This leads to the score estimate with classifier-free guidance (CFG~\cite{ho2022classifier}):
\begin{equation}
\begin{split}
\epsilon_\phi^{\textbf{CFG}}(\mathbf{z_{t}},c_I,c_P) = \; & \alpha_1[\epsilon_\phi(\mathbf{z_{t}},c_I,c_P)-\epsilon_\phi(\mathbf{z_{t}},c_P,\varnothing)] + \\ 
& \alpha_2[\epsilon_\phi(\mathbf{z_{t}},c_P,\varnothing) -\epsilon_\phi(\mathbf{z_{t}},\varnothing,\varnothing)] + \\ & \epsilon_\phi(\mathbf{z_{t}},\varnothing,\varnothing)
\end{split}
\end{equation}
\noindent where $\epsilon_{\phi}$ is a neural network for predicting noises given conditions, $\alpha_1$ and $\alpha_2$ are guidance scales that enable separately trading off the strength of conditions $c_I$ and $c_P$ separately. To ensure accurate prediction of $\epsilon_\phi(\mathbf{z_{t}}, c_P, \varnothing)$ and $\epsilon_\phi(\mathbf{z_{t}},\varnothing, \varnothing)$, one needs to randomly drop $c_I$ and $c_P$ during training $\epsilon_\phi$. However, we observed that Zero-1-to-3 did not truly follow this procedure to supervise unconditional noise. As shown in Fig.~\ref{Zero123}, Zero-1-to-3 only randomly dropped image conditions $c_I$ while keeping $c_P$ untouched. During inference, it replaces the tensor after fully connected (FC) layer $f(\cdot)$ with zeros (not exactly equal to setting both $c_I$ and $c_P$ to zeros as $f(0) \neq 0$), 
%(actually, not equal to setting both $c_I$ and $c_P$ to zeros, $f(0) \neq 0$) for unconditional noise, which 
leads to a bias. Moreover, the dataset used for fine-tuning is majorly object-centric, which may also introduce additional domain bias. Fully addressing this bias problem requires re-training the multi-view diffuser on a broader and more balanced dataset, which requires tremendous effort. Here we propose an effortless fix in the following.

\noindent \textbf{Rectification.} To alleviate the inaccurate unconditional noise problem, we first set $\alpha_1=\alpha_2=\omega$ to eliminate $\epsilon_\phi(\mathbf{z_{t}}, c_P, \varnothing)$. We then  and yield a special case of $\epsilon_\phi^{\textbf{CFG}}(\mathbf{z_{t}},c_I,c_P)$  as:
\begin{equation}
\begin{split}
\epsilon_\phi^{\textbf{CFG}}(\mathbf{z_{t}},c_I,c_P) = \; & \omega[\epsilon_\phi(\mathbf{z_{t}},c_I,c_P)-\epsilon_\phi(\mathbf{z_{t}},\varnothing,\varnothing)] + \\
& \epsilon_\phi(\mathbf{z_{t}},\varnothing,\varnothing)
\end{split}
\end{equation}

\noindent To demonstrate our setting of $\alpha_1$ and $\alpha_2$ is valid, we provide an empirical validation can be found in the Appendix \textcolor{red}{E}. Further, we consider $\epsilon_\phi(\mathbf{z_{t}},\varnothing, \varnothing)$ predicts noises from only the noisy latent and is equivalent to the unconditional noise $\epsilon_{\psi}\left(\mathbf{z_{t}}, \varnothing \right)$ of Stable Diffusion (SD) as it uses the same variational autoencoders (VAE~\cite{kingma2013auto}) as Zero-1-to-3. As such, the bias has been well rectified. To verify the effect of our rectification, Tab.~\ref{denoising} shows the denoising effect using various unconditional noise settings: (1) use SD unconditional noise $\epsilon_{\psi}\left(\mathbf{z_{t}}, \varnothing\right)$. (2) use Zero-1-to-3 noise $\epsilon_{\phi}\left(\mathbf{z_{t}}, f(0, c_P)\right)$ with $f(0, c_P)$ as a condition (same to the training process as in Fig~\ref{Zero123}). (3) use Zero-1-to-3 noise $\epsilon_{\phi}\left(\mathbf{z_{t}}, 0\right)$ with $0$ as a condition  (referring to the inference process, as shown in Fig~\ref{Zero123}). The SD noise generates the best result, and setting (2) is slightly better than (3) as it follows the training set up. More details can be found in the Appendix \textcolor{red}{A}.

\begin{table*}[t]
\centering
\scalebox{0.87}{
\begin{tabular}{lccc|ccc|ccc|ccc}
\\ \hline
Noise level & \multicolumn{3}{c}{\textbf{T}=50} & \multicolumn{3}{|c}{\textbf{T}=100} & \multicolumn{3}{|c}{\textbf{T}=200} & \multicolumn{3}{|c}{\textbf{T}=300}\\
\hline
Metrics & \multicolumn{1}{c}{PSNR$\uparrow$} & \multicolumn{1}{c}{SSIM$\uparrow$} & \multicolumn{1}{c}{LPIPS$\downarrow$} & \multicolumn{1}{|c}{PSNR$\uparrow$} & \multicolumn{1}{c}{SSIM$\uparrow$} & \multicolumn{1}{c}{LPIPS$\downarrow$} & \multicolumn{1}{|c}{PSNR$\uparrow$} & \multicolumn{1}{c}{SSIM$\uparrow$} & \multicolumn{1}{c}{LPIPS$\downarrow$} & \multicolumn{1}{|c}{PSNR$\uparrow$} & \multicolumn{1}{c}{SSIM$\uparrow$} & \multicolumn{1}{c}{LPIPS$\downarrow$}
\\ \hline                                                 
$\epsilon_\psi(\mathbf{z_{t}},\varnothing)$ & \cellcolor[RGB]{255, 153, 153}40.26  & \cellcolor[RGB]{255, 153, 153}0.992  & \cellcolor[RGB]{255, 153, 153}0.008  & \cellcolor[RGB]{255, 153, 153}36.94  & \cellcolor[RGB]{255, 153, 153}0.988  & \cellcolor[RGB]{255, 153, 153}0.014  & \cellcolor[RGB]{255, 153, 153}33.99  & \cellcolor[RGB]{255, 153, 153}0.981  & \cellcolor[RGB]{255, 153, 153}0.023  & \cellcolor[RGB]{255, 153, 153}31.98  & \cellcolor[RGB]{255, 153, 153}0.974  & \cellcolor[RGB]{255, 153, 153}0.027 \\
$\epsilon_\phi(\mathbf{z_{t}},f(0, c_P))$           & \cellcolor[RGB]{255, 204, 153}36.17  & \cellcolor[RGB]{255, 204, 153}0.981  & \cellcolor[RGB]{255, 204, 153}0.015  & \cellcolor[RGB]{255, 204, 153}34.85  & \cellcolor[RGB]{255, 204, 153}0.979  \cellcolor[RGB]{255, 204, 153}& \cellcolor[RGB]{255, 204, 153}0.024 & \cellcolor[RGB]{255, 204, 153}33.07  & \cellcolor[RGB]{255, 204, 153}0.975  & \cellcolor[RGB]{255, 204, 153}0.031  & \cellcolor[RGB]{255, 204, 153}30.44  & \cellcolor[RGB]{255, 204, 153}0.969  & \cellcolor[RGB]{255, 204, 153}0.046\\ 
$\epsilon_\phi(\mathbf{z_{t}},0)$  & \cellcolor[RGB]{255, 248, 174}36.13  & \cellcolor[RGB]{255, 248, 174}0.979  & \cellcolor[RGB]{255, 204, 153}0.015  & \cellcolor[RGB]{255, 248, 174}34.70  & \cellcolor[RGB]{255, 248, 174}0.970  & \cellcolor[RGB]{255, 248, 174}0.026  & \cellcolor[RGB]{255, 248, 174}32.76  & \cellcolor[RGB]{255, 248, 174}0.973  & \cellcolor[RGB]{255, 248, 174}0.033  & \cellcolor[RGB]{255, 248, 174}29.94  & \cellcolor[RGB]{255, 248, 174}0.963  & \cellcolor[RGB]{255, 248, 174}0.055\\ 
\hline
\end{tabular}
}
\caption{
Quantitative evaluation of different unconditional noises. We randomly samples 1,000 different images from the GSO dataset, RTMV, and Objaverse, add a certain level of noise w.r.t. \textbf{T}, where larger \textbf{T} means more noise is added, and compare the denoising effect under different conditions of Stable Diffusion and Zero-1-to-3. The top three for each metric are highlighted in \textcolor{rred}{red} , \textcolor{oorange}{orange}, and \textcolor{ydyellow}{yellow} respectively.
}
\label{denoising}
\vspace{-8pt}
\end{table*}

\noindent \textbf{Unbiased Score Distillation.} One major application of a multi-view diffuser is to distill 3D content, represented as NeRF paramater $\theta$, using the SDS loss for optimization:

\begin{equation}
\nabla_\theta \mathcal{L}_{\text{SDS}}= \mathbb{E}_{t,c_P,\epsilon }\left [w(t)\left( \epsilon_\phi^{\textbf{CFG}}\left(\mathbf{z_{t}},c_I,c_P\right) - \epsilon  \right) \dfrac{\partial\mathbf{z_{t}}}{\partial\theta}\right ]
\label{SDS}
\end{equation}

\noindent where $w(t)$ is a weighting function, $\epsilon$ is standard Gaussian noise, and $\mathbf{z_{t}}$ refers to the noisy latent as $\mathbf{z_{t}} = \sqrt{\alpha_t} \mathbf{z} + \sqrt{1-\alpha_t} \epsilon$, with $\alpha_t$ being the noise scheduler. We rewrite the noise difference in Formula \ref{SDS} by adding an additional weighting factor $\boldsymbol{\lambda}$ to $\left[\epsilon_{\psi}\left(\mathbf{z_{t}}, \varnothing \right)- \epsilon \right]$:

\begin{equation}
\begin{split}
\epsilon_\phi^{\textbf{CFG}}(\mathbf{z_{t}},c_I,c_P) - \epsilon = \;& \omega \left[ \epsilon_{\phi}\left(\mathbf{z_{t}}, c_I,c_P\right)-\epsilon_{\psi}\left(\mathbf{z_{t}}, \varnothing \right) \right] + \\
& \boldsymbol{\lambda} \left[\epsilon_{\psi}\left(\mathbf{z_{t}}, \varnothing \right)- \epsilon \right]
\end{split}
\end{equation}

\noindent where setting $\boldsymbol{\lambda}=1$ we get Formula \ref{SDS}. Inspired by DDS ~\cite{hertz2023delta} and CSD~\cite{yu2023text}, we observed that setting $\boldsymbol{\lambda}=0$ can significantly improve the details of the 3D details generated using SDS. Further details and an in-depth analysis are provided in the Appendix \textcolor{red}{D}. We obtain our Unbiased Score Distillation (USD) as:
\begin{equation}
\nabla_\theta \mathcal{L}_{\text{USD}} = \mathbb{E}_{t, c_P, \epsilon }\left [w(t) \left[ \omega \left( \epsilon_{\phi}\left(\mathbf{z_{t}}, c_I,c_P\right)- \epsilon_{\psi}\left(\mathbf{z_{t}}, \varnothing \right) \right) \right] \dfrac{\partial\mathbf{z_{t}}}{\partial\theta}\right ]
\label{UMSDS}
\end{equation}
\noindent Our USD generates much better and consistent 3D than the SDS/SJC method used in Zero-1-to-3, the results can be found in the Appendix \textcolor{red}{E}.

\subsection{Consistent View and Geometry Distillation}
\label{sec:Consistent View and Geometry Distillation}
Although USD can be used to optimize NeRF for view synthesis, the resulting images still tend to be blurry, and directly extracting geometry from the NeRF density field introduces noise into the mesh. Since our ultimate goal is to extract high-quality geometry and consistent multi-views, we propose to utilize the generated NeRF as the view consistency prior, \textit{i.e.}, the final high-quality view images should be consistent with the NeRF rendering. Thus, our problem transformed into a denoising problem. 
 
\noindent \textbf{Two-Stage Specified Diffusion.} We leverage the recent advance in diffusion model specialization, \textit{i.e.}, the DreamBooth~\cite{ruiz2023dreambooth}, to fine-tune a 2D diffusion model for the specific target in the input image. We observe that the novel view images generated by Zero-1-to-3, though not multi-view consistent, tend to have the same style as the input image. We design a two-stage tuning method to gradually let the diffusion model learn the object details. In the first stage, we use the multi-view images generated by Zero-1-to-3 as positive samples, contrasting them with text-prompt generated images as negative samples for learning the visual style of the target. In the subsequent stage, the input image serves as the positive sample, with all Zero-1-to-3 generated images as negatives, focusing on learning finer details. During optimization, we use a unique identifier {\color{ggrean} \textbf{[V]}} to capture the visual style of the target. In the second stage, we set only the input image as the positive sample and the images generated by Zero-1-to-3 as negative ones, and use the additional identifier {\color{red} \textbf{[S]}} to capture the identity of the specific target.

\noindent \textbf{Geometry and Texture Distillation with Input View Supervision.} With the specialization diffusion model, we add a small noise, Stable Diffusion scheduler $t=200$, to the NeRF render images and conduct the denoising process. Then, we use the NeuS technique to reconstruct the geometry from the high-quality and clear images (\textit{i.e.}, 100 input images). 

We observe that the input view is rarely used for optimization of the geometry and texture in multi-view diffusers, despite the fact that the input image is most faithful to the target object. To exploit the input view information, we further develop a reference view score distillation during the NeuS reconstruction process. Specifically, we consider the image render from NeuS at the input image viewpoint as $\mathbf{z_{t}}^* = \mathcal{R}(\Theta, p^*)$, where $\mathcal{R}$ is the rendering function from NeuS model defined by $\Theta$, $p^*$ is the camera pose of the input image, can be set to a particular relative pose. We define the following reference view distillation loss as:
\begin{equation}
\begin{aligned}
\mathcal{L}_{RV} = \mathbb{E}_{t, \epsilon}\left [ w(t) || \epsilon_{\psi} \left(\mathbf{z_{t}}^*, \varnothing\right)-\epsilon_{\psi} \left(\mathbf{y_{t}}, \varnothing\right) ||_{2}^{2} \right ]
\end{aligned}
\end{equation}
where $\mathbf{y_{t}}$ is input view image. We add this reference view distillation loss $\mathcal{L}_{RV}$ to original photometric loss with optimized images for optimizing the NeuS parameter $\Theta$. This scheme is a better supervision strategy than directly applying MSE loss on the input view. The reason for this is that, compared to computing MSE loss directly at the pixel-level, our patch-aware noise(latent)-level approach places greater emphasis on the perceptual quality of the image.

\begin{figure*}[t]
    \centering
    \includegraphics[width=0.91 \linewidth]{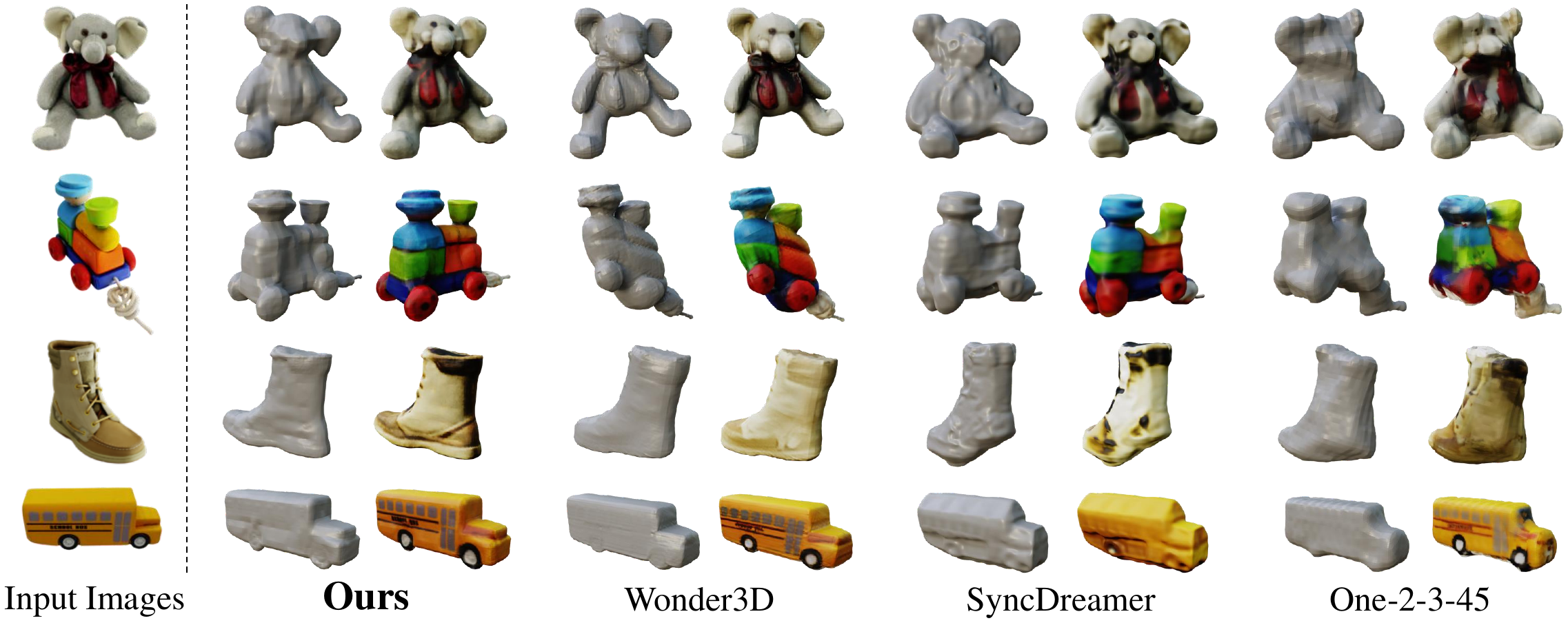}
    \caption{Qualitative comparisons of our method with baseline approaches, namely Wonder3D, SyncDreamer, and One-2-3-45, on the GSO dataset, focusing on the quality of the reconstructed textured meshes.}
    \label{meshcompare}
    \vspace{-8pt}
\end{figure*}

\section{Experiments}
We conduct extensive experiments, both qualitatively and quantitatively, to demonstrate the effectiveness of our method. 

\subsection{Implementation Details}

\noindent \textbf{Optimizing the Pipeline.}  We use exactly the same set of hyperparameters for all experiments and do not perform any per-object hyperparameter optimization. We use the implicit-volume implementation in threestudio~\cite{threestudio2023} as our 3D representation (NeRF), which includes a multi-resolution hash-grid and a MLP to predict density and RGB. The NeRF is optimized for 10,000 steps with an Adam optimizer at a learning rate of 0.01, weight decay of 0.05, and betas of (0.9, 0.95). For USD, the maximum and minimum time steps are decreased from 0.98 to 0.5 and 0.02, respectively, over the first 5,000 steps. We adopt the Stable Diffusion~\cite{takagi2023high} model of V2.1.  The classifier-free guidance (CFG) scale of the USD is set to 7.5 following ~\cite{wang2023prolificdreamer}. The DreamBooth backbone is implemented using Stable Diffusion V2.1. In the first stage, we use Stable Diffusion to generate 200 images as negative samples. Additionally, we utilize 6 positive sample images with 360$^{\circ}$ surrounding camera poses (at 60$^{\circ}$ intervals) for training. The USD (NeRF) process takes about 1.5 hours on a NVIDIA Tesla V100 (32GB) GPU. To achieve reduced running time, we provide additional discussions and experimental results in the Appendix \textcolor{red}{C}. For DreamBooth fine-tuning, we train the model around for 600 steps with a learning-rate as 2e-6, weight decay as 0.01 and a batch size of 2. More details on the experimental setup are provided in the Appendix \textcolor{red}{B}. 

\noindent \textbf{Camera Setting.} Since the reference image is unposed, we assume its camera parameters ~\cite{hugging2023one2345} are as follows. We set the field of view (FOV) of the camera is 40$^{\circ}$, and the radial distance is 1.5 meters. Note this camera setting works for images subject to the front-view assumption. For images taken deviating from the front view, a manual change of polar angle or a camera estimation is required. 

\subsection{Evaluation Protocol}
\noindent \textbf{Evaluation Datasets.} Following prior research~\cite{liu2023zero,liu2023syncdreamer,long2023wonder3d}, we adopt the Google Scanned Object dataset~\cite{downs2022google} for our evaluation, which includes a wide variety of common everyday objects. 
Our evaluation dataset matches that of SyncDreamer~\cite{liu2023syncdreamer}, consisting of 30 objects that span from everyday items to animals.
For each object in the evaluation set, we render an image with a size of 256 $\times$ 256 and use it as the input. Additionally, to assess the generalization ability of our model, we include images with diverse styles collected from the website in Zero-1-to-3, SyncDreamer and Wonder3D.

\noindent

\begin{table}[t]
\centering
\scalebox{0.77}{
\begin{tabular}{lcc}
   \toprule
   Method  & Chamfer Dist.$\downarrow$ & Volume IoU$\uparrow$ \\
   \midrule
   Realfusion~\cite{melas2023realfusion}    
   & 0.0819  & 0.2741   \\
   Magic123~\cite{qian2023magic123}
   & 0.0516 &  0.4528 \\
   One-2-3-45~\cite{liu2023one}    
   & 0.0629 &  0.4086 \\
   Point-E~\cite{nichol2022point}    
   & 0.0426 & 0.2875 \\
   Shap-E~\cite{jun2023shap}    
   & 0.0436 &  0.3584  \\
   Zero-1-to-3~\cite{liu2023zero}    
   & 0.0339 &  0.5035 \\
   SyncDreamer~\cite{liu2023syncdreamer} \tiny{(NeuS)}
   &  0.0261  &  0.5421  \\
   Wonder3D~\cite{long2023wonder3d} \tiny{(iNGP+NeuS)}
   &  \cellcolor[RGB]{255, 204, 153}0.0199  &  \cellcolor[RGB]{255, 204, 153}0.6244   \\
   % \midrule
   \hline           
   Ours \tiny{(NeuS)}
   &  \cellcolor[RGB]{255, 248, 174}0.0240  &  \cellcolor[RGB]{255, 248, 174}0.5688  \\
   Ours \tiny{(iNGP+NeuS)}
   &  \cellcolor[RGB]{255, 153, 153}0.0177  &  \cellcolor[RGB]{255, 153, 153}0.6330  \\
   \bottomrule
\end{tabular}
}
\caption{Quantitative comparison with baseline methods. We report Chamfer Distance and Volume IoU on the GSO dataset. The original implementation of SyncDreamer uses vanilla NeuS for extracting 3D meshes, while Wonder3D uses Instant NGP (iNGP) adapted NeuS. We report results using both techniques for better demonstration.
}
\label{tab:recon}
% \vspace{-8pt}
\end{table}

\begin{table}[t]
\centering
\scalebox{0.77}{
\begin{tabular}{lccc}
   \toprule
   Method  & PSNR$\uparrow$ & SSIM$\uparrow$ & LPIPS$\downarrow$  \\
   \midrule
   % DietNeRF~\cite{jain2021putting}    
   % &       &       &       &      \\
   Realfusion~\cite{melas2023realfusion}    
   & 15.26 & 0.722 & 0.283   \\
   Zero-1-to-3~\cite{liu2023zero}    
   & 18.93 & 0.779 & 0.166   \\
   SyncDreamer~\cite{liu2023syncdreamer}   
   & \cellcolor[RGB]{255, 248, 174}20.05 & \cellcolor[RGB]{255, 248, 174}0.798 & \cellcolor[RGB]{255, 248, 174}0.146 \\
   Wonder3D~\cite{long2023wonder3d}
   & \cellcolor[RGB]{255, 153, 153}26.07 & \cellcolor[RGB]{255, 204, 153}0.924 & \cellcolor[RGB]{255, 204, 153}0.065 \\
   % \midrule
   \hline           
   Ours    
   % & \cellcolor[RGB]{255, 204, 153}23.62 & \cellcolor[RGB]{255, 153, 153}0.931 & \cellcolor[RGB]{255, 204, 153}0.075 \\
   & \cellcolor[RGB]{255, 204, 153}25.38 & \cellcolor[RGB]{255, 153, 153}0.927 & \cellcolor[RGB]{255, 153, 153}0.049 \\
   \bottomrule
\end{tabular}
}
\caption{The quantitative comparison in novel view synthesis. We report PSNR, SSIM, LPIPS on the GSO dataset.}
\label{tab:nvs}
\vspace{-8pt}
\end{table}

\begin{table}[t]
\centering
\scalebox{0.8}{
\begin{tabular}{lcc}
   \toprule
   Method  & Chamfer Dist.$\downarrow$ & Volume IoU$\uparrow$ \\
   \midrule
   % \midrule        
   w/o USD
   &  0.0253  &  0.5515  \\
   w/o DB[1st+2nd]
   &  0.0217  &  0.6023  \\
   w/o DB[2nd]
   &  0.0190  &  0.6216  \\
   w/o $\mathcal{L_{\textit{RV}}}$
   &  0.0185  &  0.6229  \\
    \midrule
   \bf Ours
   &  \bf0.0177  &  \bf0.6330  \\
   \bottomrule
\end{tabular}
}
\caption{Quantitative results of ablation studies. We report Chamfer Distance and Volume IoU on the GSO dataset.}
\label{tab:mesh}
\vspace{-8pt}
\end{table}

\noindent \textbf{Metrics.} To evaluate the quality of single view reconstruction, we used two commonly used metrics: the chamfer distance (CD) between the ground truth shape and the reconstructed shape, and the volume IoU.
Due to different methods using different normative systems, before calculating these two metrics, we first align the generated shapes with the basic fact shapes.
Moreover, we adopt the metrics PSNR, SSIM~\cite{wang2004image}, LPIPS~\cite{zhang2018unreasonable} for evaluating the generated color images.

\subsection{Ablation Study}

\begin{figure}[t]
    \centering
    \includegraphics[width=1.0\linewidth]{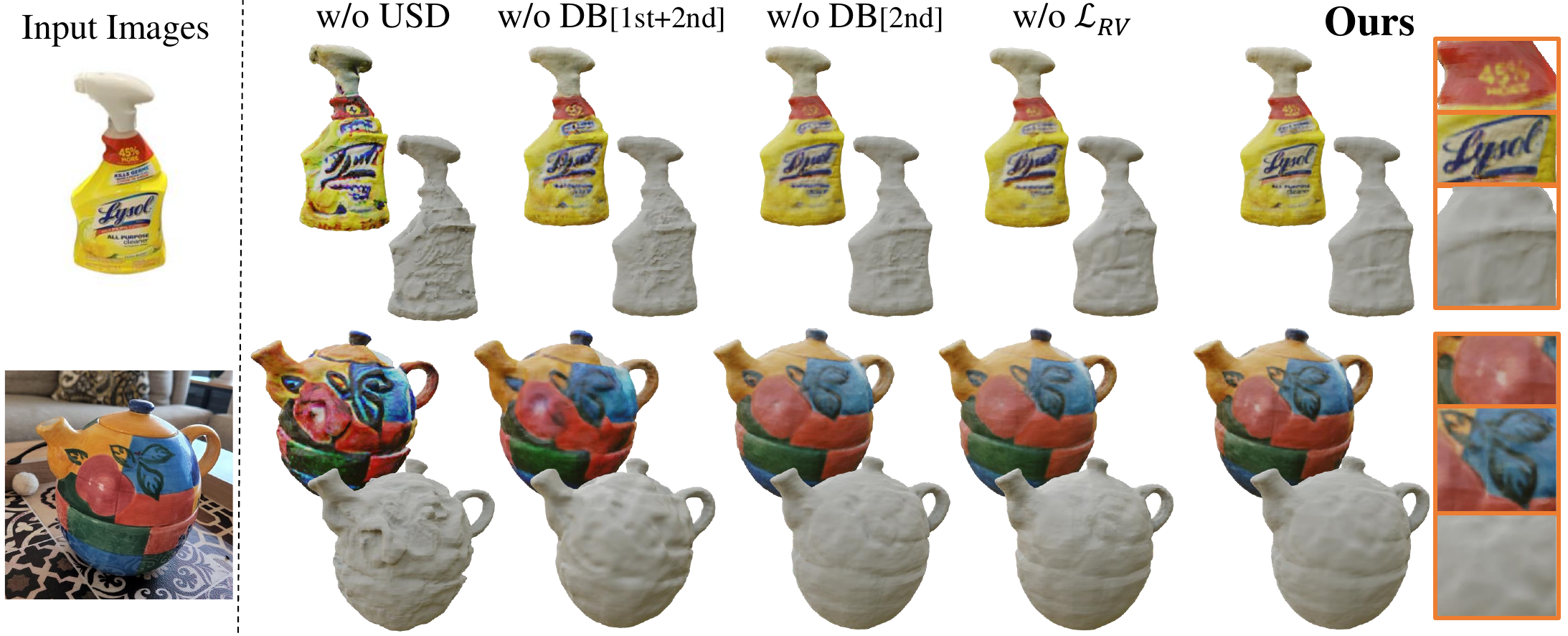}
    \caption{Ablation study on the effect of each components, including the USD (Sec.~\ref{sec:Unbiased Multi-view Score Distillation}), the reference view score distillation (Sec.~\ref{sec:Consistent View and Geometry Distillation}), the DreamBooth (Sec.~\ref{sec:Consistent View and Geometry Distillation}) refinement of our method.
    }
    \label{ablation}
    \vspace{-8pt}
\end{figure}

We validate our design choices by ablating 4 major model variants, that are without Unbiased Score Distillation, without reference view distillation loss, without the two-stage DreamBooth (DB). As shown in Fig.~\ref{ablation}, the 3D models generated without USD exhibit biased texture colors, and their shapes are not smooth. Not using reference view supervision will result in the inability to recover the same texture details as the input image, and not using DB will result in blurring of texture details. We also conducted a quantitative analysis on the GSO dataset, presented in Tab.~\ref{tab:mesh}. The results demonstrate that USD is crucial for geometric accuracy, and all our sub-modules collectively enhance overall performance.

\subsection{Different Viewing Angle Comparisons}
We found that SyncDreamer~\cite{liu2023syncdreamer} and Wonder3D~\cite{long2023wonder3d} are very sensitive to viewing angles. If a relatively high viewing angle is input, SyncDreamer and Wonder3D will predict incorrect multi-view images. The results can be found in the Appendix \textcolor{red}{E}.

\subsection{Single View Reconstruction}
We evaluate the quality of the reconstructed geometry of different methods. The quantitative results are summarized in Tab.~\ref{tab:recon}, and the qualitative comparisons are presented in Fig.~\ref{meshcompare}. The quality of Wonder3D shape reconstruction depends on the perspective of the input view, such as `$\textit{Train}$' shown in Fig.~\ref{meshcompare}, where Wonder3D generated incorrect prediction results. The shape generated by SyncDreamer undergoes deformation due to the camera pose in the input view. One-2-3-45~\cite{liu2023one} attempts to reconstruct meshes from the multiview-inconsistent outputs of Zero-1-to-3~\cite{liu2023zero}. While it can capture coarse geometries, it loses important details in the process. In contrast, our method can achieve good reconstruction quality and texture in terms of geometric structure and texture. In the paper, all the surface extraction demonstrated by our method is built on the Instant NGP (iNGP)~\cite{muller2022instant} based SDF reconstruction method~\cite{instant-nsr-pl}, and we use Blender Cycles~\cite{blender} to render the results.

\subsection{Novel View Synthesis}
We evaluate the quality of novel view synthesis for different methods. The quantitative results are presented in Tab.~\ref{tab:nvs}, and the qualitative results can be found in Fig.~\ref{color-compare}. 
Zero-1-to-3 produces visually reasonable images, but they lack multi-view consistency since it operates on each view independently (we don't show the results of Zero-1-to-3).  Although SyncDreamer introduces a volume attention scheme to enhance the consistency of multi-view images, their model is sensitive to the elevation degrees of the input images and tends to produce unreasonable results. Wonder3D ensures 3D consistency by generating normal maps, but may result in incorrect results for some input viewpoints.

\subsection{Text-to-Image-to-3D}
As a case study, we combine text-to-image models, \textit{i.e.}, the Stable Diffusion or Imagen ~\cite{saharia2022photorealistic} to generate 3D models from text. We show some examples in the Appendix \textcolor{red}{E}.
%Fig.~\ref{fig:Textto3D}. 
Compared to DreamFusion~\cite{poole2022dreamfusion}, ProlificDreamer~\cite{wang2023prolificdreamer} and MVDream~\cite{MVDream}, our method shows no multi-face Janus problem and conforms to the text faithfully.

\section{Conclusions}
In this work, we introduce an optimized approach for distilling geometry and views from a multi-view diffuser, with a specific focus on the Zero-1-to-3 model. We observed that the direct application of the SDS/SJC technique to Zero-1-to-3 is often suboptimal, primarily due to bias issues inherent in unconditional noise. To address this, we propose an Unbiased Score Distillation (USD) strategy by leveraging unconditioned noises from a 2D diffusion model to effectively enhance the optimized radiance field.
Moreover, we developed a two-stage DreamBooth refinement process to improve the rendering of views. This process ensures consistency across multiple perspectives while simultaneously enhancing image quality.
While we have identified and addressed the bias issue in the Zero-1-to-3 model, the underlying causes remain to be fully understood. Future research will delve into the theoretical aspects of this bias problem. Additionally, we aim to explore the potential applications of USD in other fields, such as image translation and view synthesis. 

\section*{Acknowledgements}
This work is supported by the National Natural Science Foundation of China (NSFC No. 62272184 and No. 62402189), the China Postdoctoral Science Foundation under Grant Number GZC20230894, the China Postdoctoral Science Foundation (Certificate Number: 2024M751012). The computation is completed in the HPC Platform of Huazhong University of Science and Technology.

\bibliographystyle{named}
\bibliography{ijcai25}

\begin{figure*}[t]
    \centering
    \includegraphics[width=1.0\linewidth]{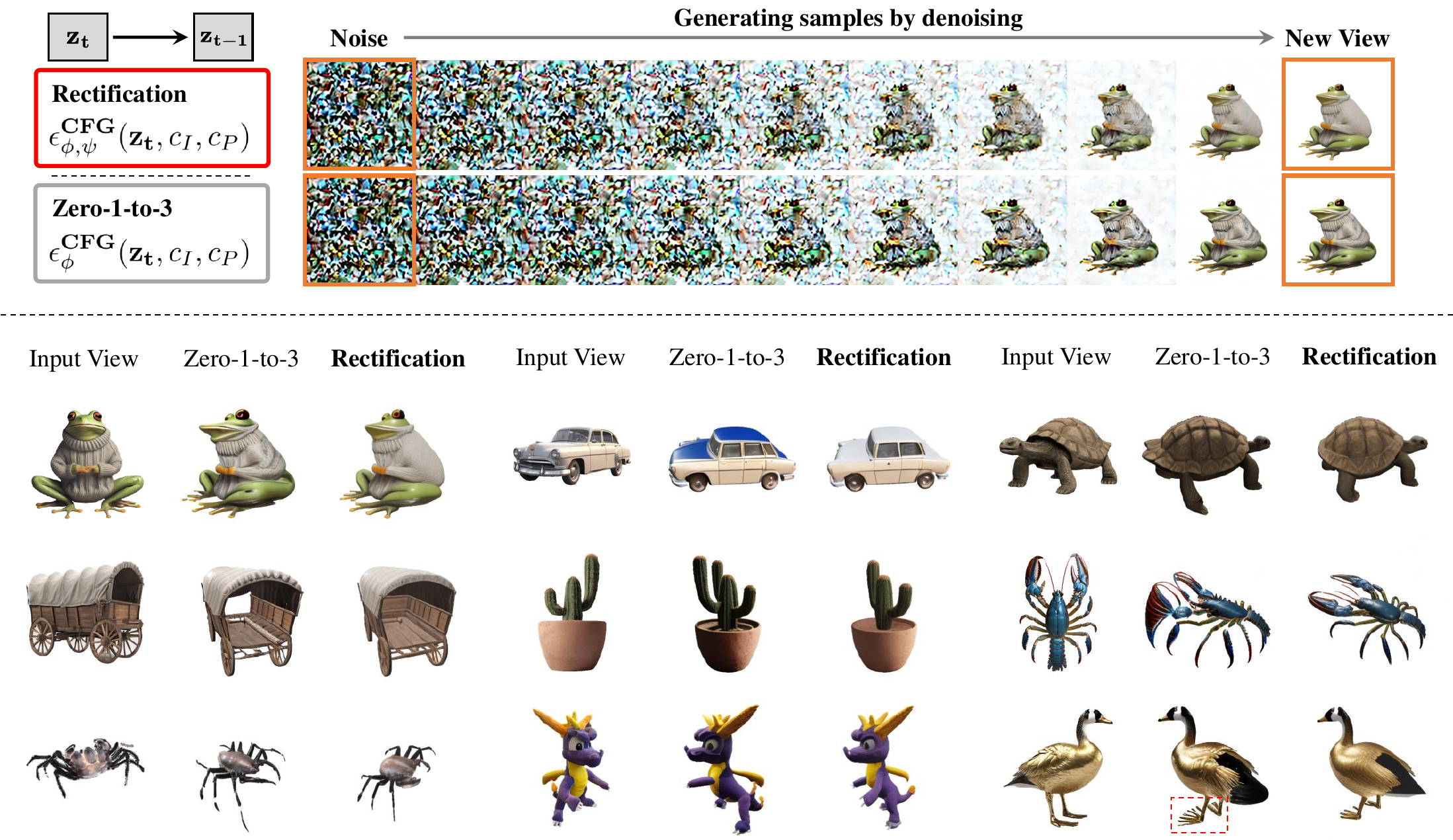}
    \caption{Novel view synthesis on in-the-wild images. Comparison between Zero-1-to-3 and our rectification. Starting from the input view, the task is to generate an image of the object under a specific camera pose transformation.
    }
    \label{fig:sampling}
\end{figure*}

\section*{Appendix}
\renewcommand\thesection{\Alph{section}}
\setcounter{section}{0}

This supplementary material provides additional information and experiment results pertaining to the main paper, “\textit{Optimized View and Geometry Distillation from Multi-view Diffuser}”, including detailed descriptions of more visual results to complement the experiments reported in the main manuscript.

\section{Unbiased Sampling of Multi-view Diffuser}

We rectify the unconditional noise in Formula $\color{red}{4}$ of the main paper:

\begin{equation*}
\resizebox{0.95\linewidth}{!}{$
\begin{split}
\begin{WithArrows}
\epsilon_\phi^{\textbf{CFG}}(\mathbf{z_{t}},c_I,c_P) & = \omega[\epsilon_\phi(\mathbf{z_{t}},c_I,c_P)-\epsilon_\phi(\mathbf{z_{t}},\varnothing,\varnothing)] + \epsilon_\phi(\mathbf{z_{t}},\varnothing,\varnothing) \Arrow[tikz=thick]{} \\
\epsilon_{\phi, \psi}^{\textbf{CFG}}(\mathbf{z_{t}},c_I,c_P) & = \omega[\epsilon_\phi(\mathbf{z_{t}},c_I,c_P)-\epsilon_{\psi}\left(\mathbf{z_{t}}, \varnothing \right) ] + \epsilon_{\psi}\left(\mathbf{z_{t}}, \varnothing \right)
\end{WithArrows}
\end{split}
$}
\end{equation*}

Our proposed rectification method essentially combines the \colorbox{intCrimLL}{unconditional noise} prediction from the base model with the \colorbox{intBlueLL}{conditional noise} prediction from the fine-tuned model.

\begin{equation*}
\mymathbox{intTealL}{intTealLL}{\epsilon_{\phi,\psi}^{\textbf{CFG}}(\mathbf{z_{t}},c)}{\textbf{Rectification}} = \omega[\mymathbox{intBlueL}{intBlueLL}{\epsilon_\phi(\mathbf{z_{t}},c)}{\textbf{Fine-tuned}}-\mymathbox{intCrimL}{intCrimLL}{\epsilon_\psi(\mathbf{z_{t}},\varnothing)}{\textbf{Base Model}}] + \mymathbox{intCrimL}{intCrimLL}{\epsilon_\psi(\mathbf{z_{t}},\varnothing)}{\textbf{Base Model}}
\end{equation*}

We found that rectifying the bias in Zero-1-to-3~\cite{liu2023zero} achieves significantly better novel views from in-the-wild inputs. Fig.~\ref{fig:sampling} shows examples from One-2-3-45++~\cite{liu2023one2345++}, Wonder3D~\cite{long2023wonder3d} and SyncDreamer~\cite{liu2023syncdreamer}. Our rectification is able to generate novel views that are more consistent with the input view. Additionally, rectification is able to generate novel views from input view while keeping the original style as well as object geometric details. These examples show the effectiveness of our rectification.

\section{Two-Stage Specified Diffusion Details}

\begin{figure}[h]
    \centering
    \includegraphics[width=1.0\linewidth]{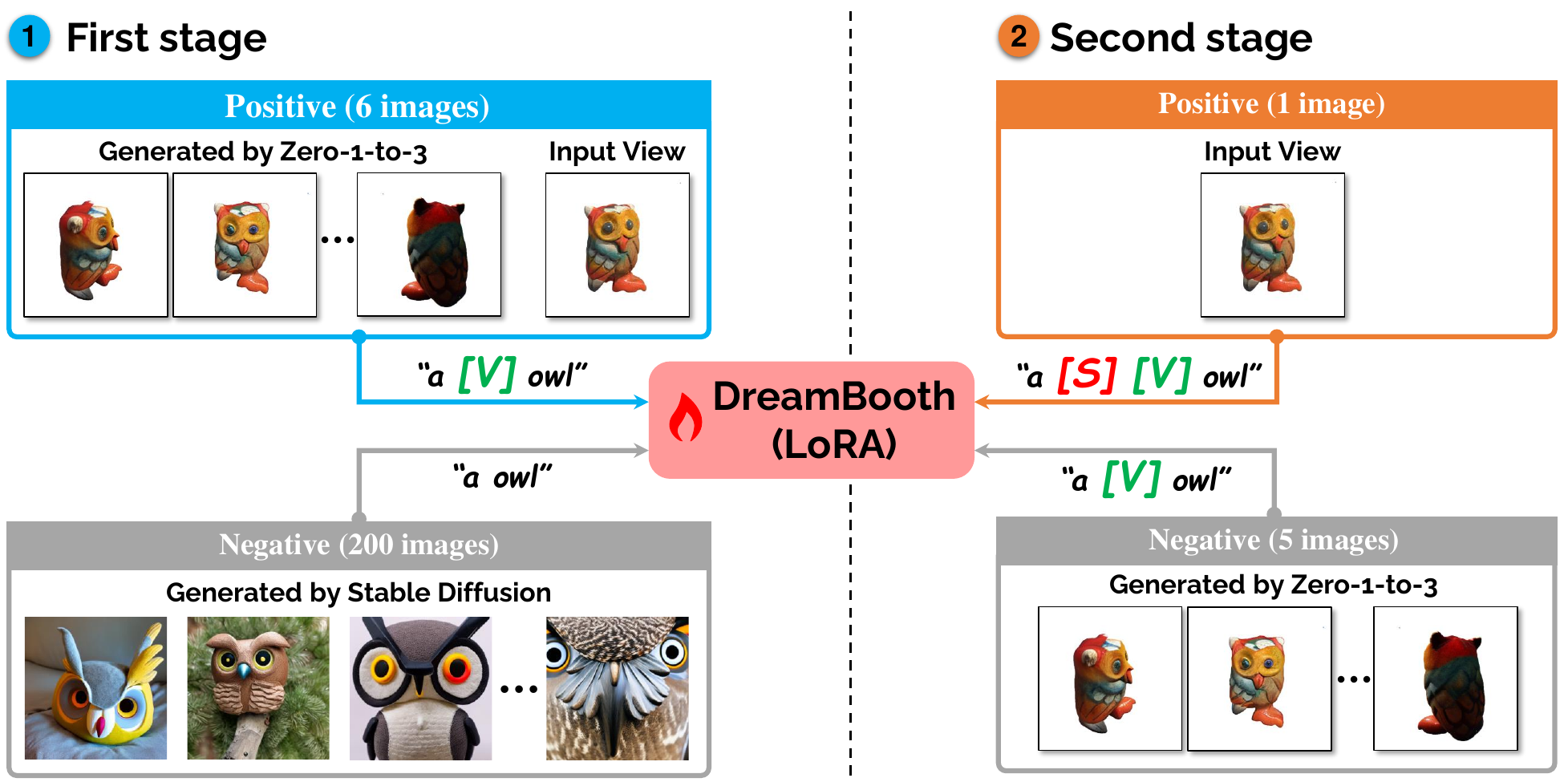}
    \caption{Fine-tuning. As a demonstration, we use the `$\textit{owl}$' image as a toy example.}
    \label{dreambooth}
\end{figure}

\begin{figure}[h]
    \centering
    \includegraphics[width=1.0\linewidth]{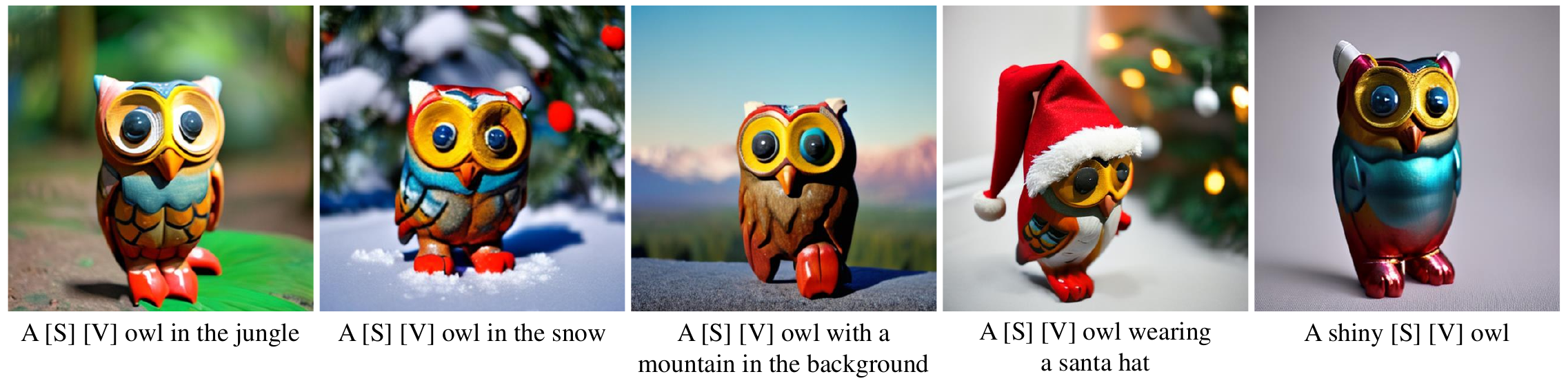}
    \caption{Results for re-contextualization of a `$\textit{owl}$' subject instances.}
    \label{fig:re-contextualization}
    \vspace{-8pt}
\end{figure}

\noindent \textbf{DreamBooth}~\cite{ruiz2023dreambooth} provides a network fine-tuning strategy to adapt a given text-to-image denoising network to generate images of a specific subject.

\noindent \textbf{Low Rank Adaptation (LoRA)}~\cite{hu2021lora} provides a memory-efficient and faster technique for DreamBooth. Priors work show that this low-rank residual fine-tuning is an effective technique that preserves several favorable properties of the original DreamBooth while also being memory-efficient as well as fast.

We implement our DreamBooth on the Stable Diffusion~\cite{rombach2022high} V2.1 diffusion model and we predict the LoRA weights for all cross and self-attention layers of the diffusion U-Net~\cite{dreamboothlora}. Fig.~\ref{dreambooth} illustrates the model fine-tuning with the class-generated samples. Fig.~\ref{fig:re-contextualization} shows a collection of generation results to illustrate how our method can generate novel images for a specific subject in different contexts with descriptive prompts. 

\begin{figure}[h]
    \centering
    \includegraphics[width=1.0\linewidth]{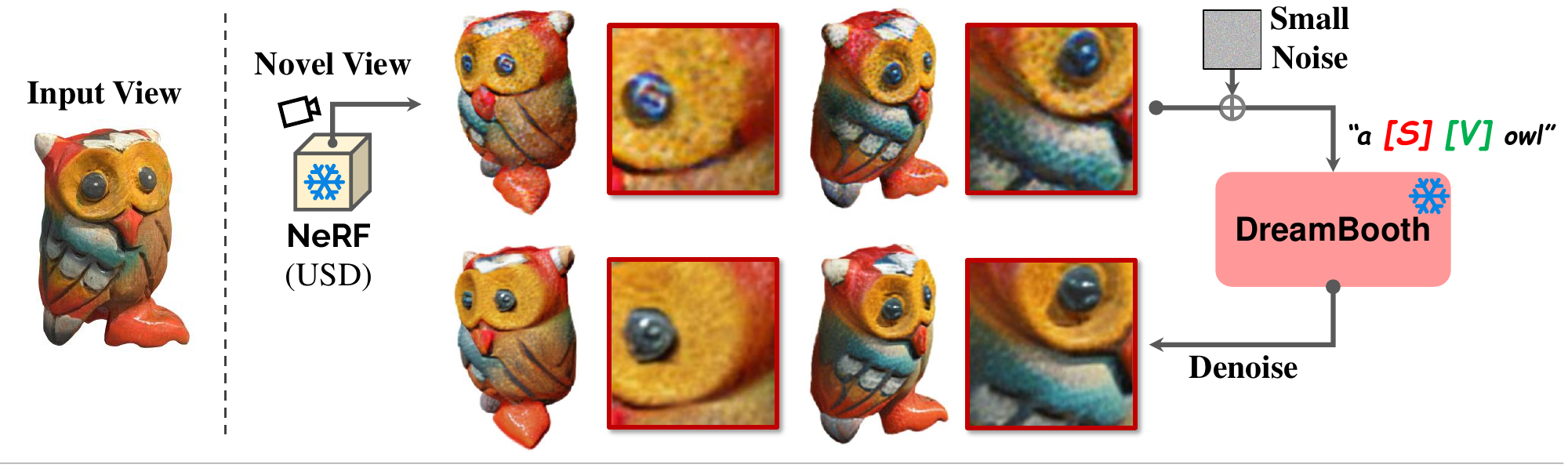}
    \includegraphics[width=1.0\linewidth]{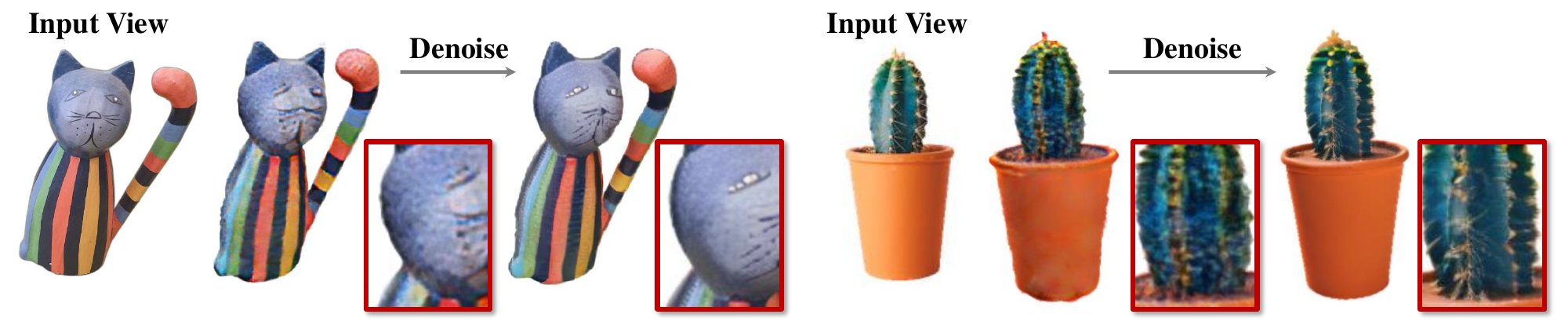}
    \caption{With the specialization diffusion model, we add a small noise, Stable Diffusion scheduler $t=200$, to the NeRF render images and conduct the denoising process.}
    \label{fig:denoise}
    \vspace{-8pt}
\end{figure}

NeRF~\cite{mildenhall2020nerf} uses a volume rendering method to learn a volumetric radiance field for novel view synthesis. However, NeRF architectures~\cite{mildenhall2020nerf,zhang2023nemf,warburg2023nerfbusters,barron2022mip} are prone to cloudy artifacts (\textit{floaters}), which it is difficult to extract a high-quality surface, as shown in Fig.~\ref{fig:denoise}. To address this issue, we denoise multi-view renderings from the trained NeRF using the fine-tuned DreamBooth model.

\section{3D Gaussian Splatting Representation}

Recently, DreamGaussian~\cite{tang2023dreamgaussian} and GaussianDreamer~\cite{yi2023gaussiandreamer} utilizes 3D Gaussians as an efficient 3D representation that supports real-time high-resolution rendering via rasterization. We adapt 3D Gaussian Splatting~\cite{kerbl20233d} into the generative setting with Unbiased Score Distillation (USD). The quantitative results are summarized in Tab.~\ref{tab:recon} and Tab.~\ref{tab:nvs}. Our method is implemented in PyTorch~\cite{paszke2019pytorch}, based on threestudio~\cite{threestudio2023}.

\begin{figure}[h]
    \centering
    \includegraphics[width=1.0\linewidth]{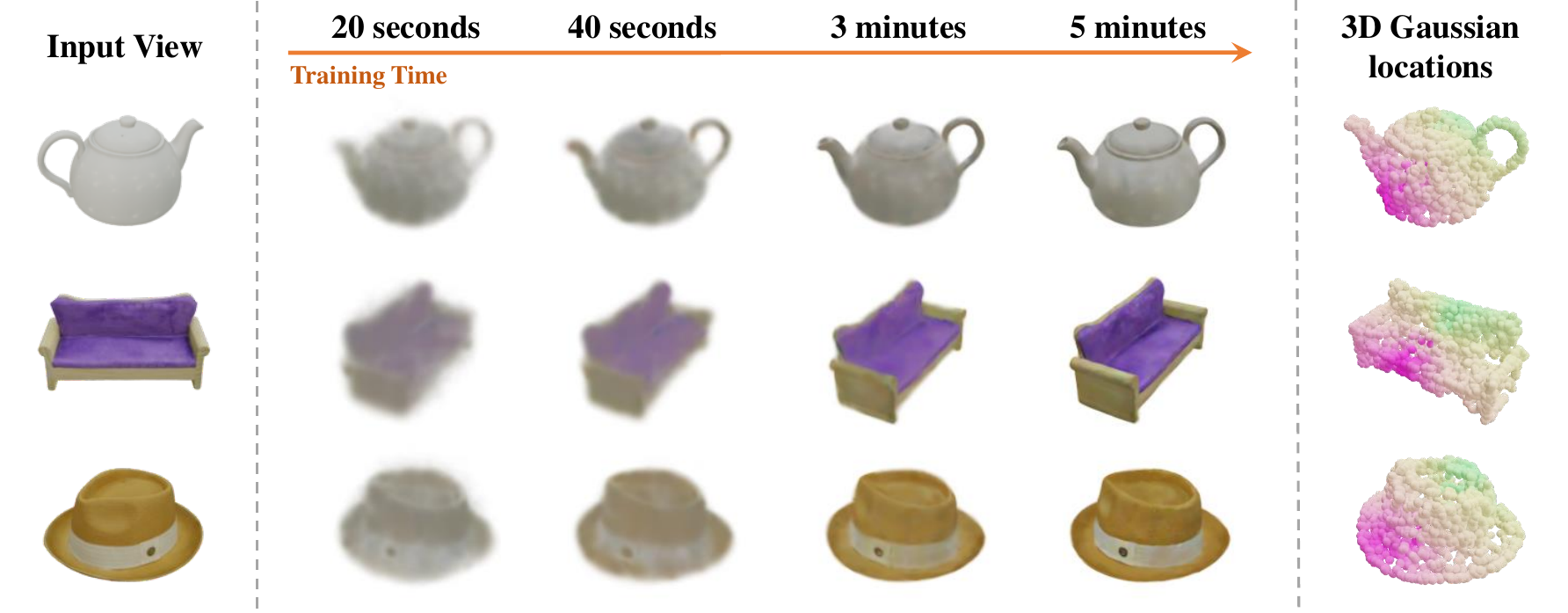}
    \caption{Optimization Progress. The shape to initializa the 3D Gaussians as sphere.}
    \label{GS}
\end{figure}

\begin{figure}[h]
    \centering
    \includegraphics[width=1.0\linewidth]{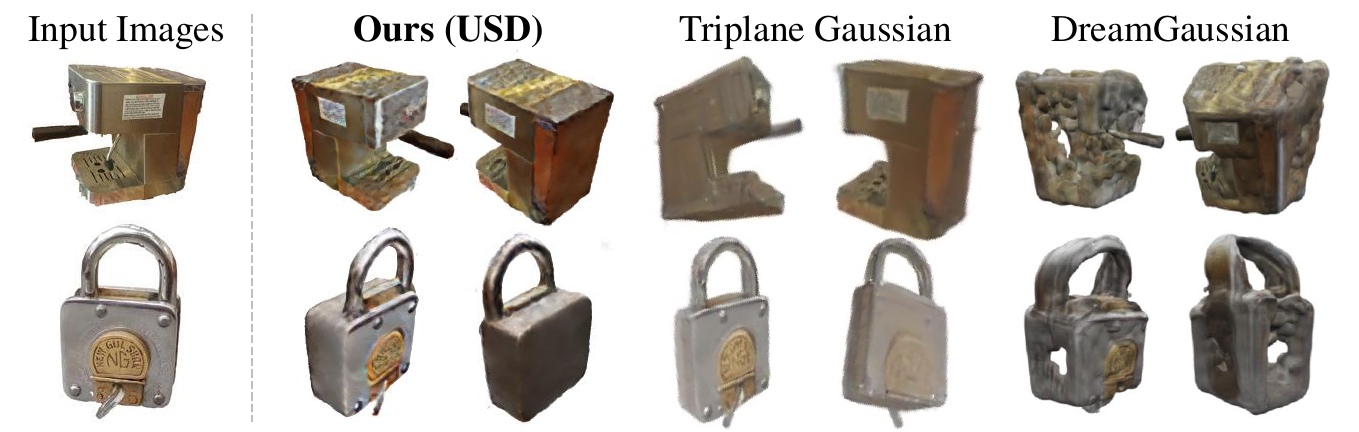}
    \caption{
    The qualitative comparisons of novel view synthesis with  DreamGaussian and Triplane Gaussian. Our approach achieves both quality and consistency across different novel views.
    }
    \label{compare}
\end{figure}

\begin{figure*}[t]
    \centering
    \includegraphics[width=1.0\linewidth]{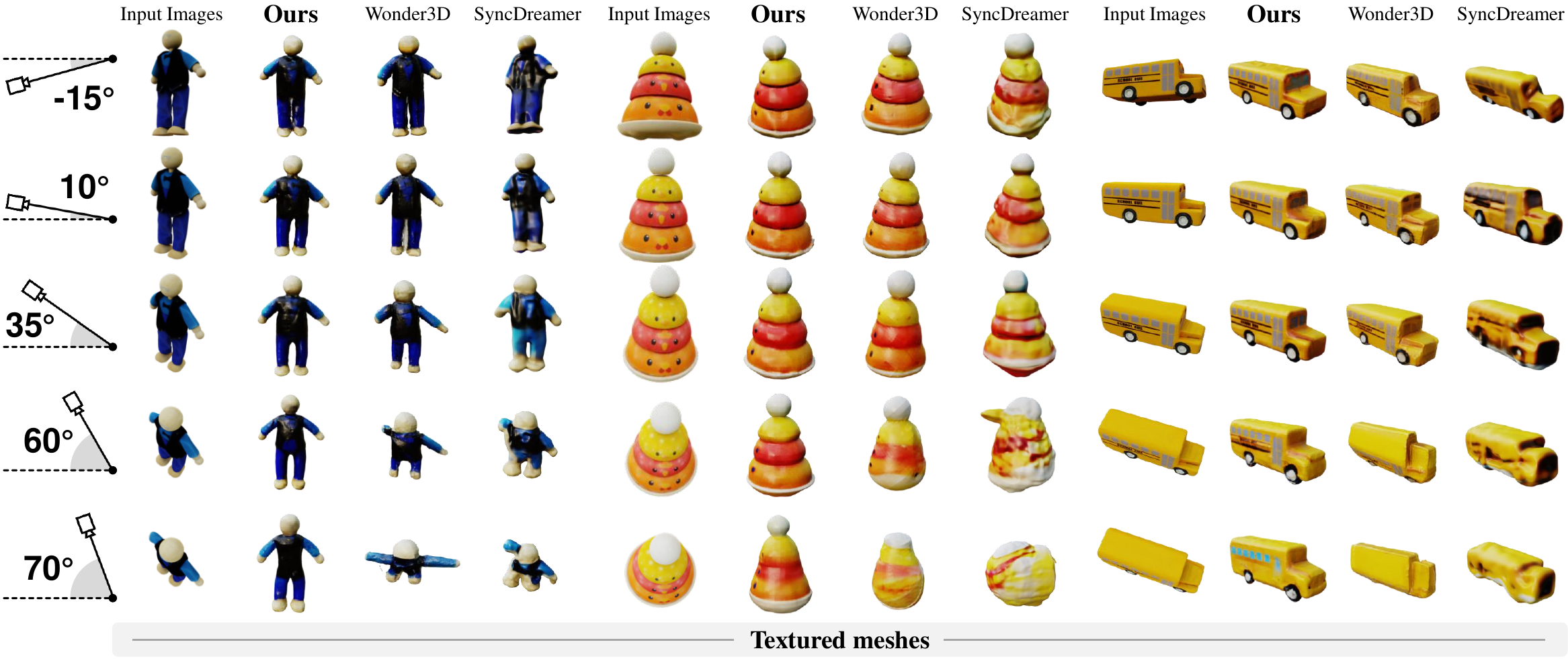}
    \caption{
        SOTA methods, like SyncDreamer and Wonder3D impose limitations on the viewing angles of the input image. For instance, these methods proficiently manage near-horizontal viewing angles but often falter with upward views. In contrast, our approach consistently generates reasonable structures across a broader range of input view poses.
    }
    \label{view_compare}
\end{figure*}

\noindent{\bf On comparison to DreamGaussian and Triplane Gaussian.}  
We here include some qualitative comparisons with DreamGaussian and Triplane Gaussian ~\cite{zou2024triplane} in Fig.~\ref{compare}.

\begin{table}[h]
% \vspace{-4pt}
\centering
\setlength{\tabcolsep}{3pt}
\caption{Quantitative comparison with baseline methods.}
\label{tab:recon}
\resizebox{0.99\linewidth}{!}{
\begin{tabular}{lcc}
   \toprule
   Method  & Chamfer Dist.$\downarrow$ & Volume IoU$\uparrow$ \\
   \midrule
   Realfusion~\cite{melas2023realfusion}    
   & 0.0819  & 0.2741   \\
   Magic123~\cite{qian2023magic123}
   & 0.0516 &  0.4528 \\
   One-2-3-45~\cite{liu2023one}    
   & 0.0629 &  0.4086 \\
   Point-E~\cite{nichol2022point}    
   & 0.0426 & 0.2875 \\
   Shap-E~\cite{jun2023shap}    
   & 0.0436 &  0.3584  \\
   Zero-1-to-3~\cite{liu2023zero}    
   & 0.0339 &  0.5035 \\
   SyncDreamer~\cite{liu2023syncdreamer} \tiny{(NeuS)}
   &  0.0261  &  0.5421  \\
   Wonder3D~\cite{long2023wonder3d} \tiny{(iNGP+NeuS)}
   &  \cellcolor[RGB]{255, 204, 153}0.0199  &  \cellcolor[RGB]{255, 204, 153}0.6244   \\
   \hline           
   3DGS+USD \tiny{(NeuS)}
   &  \cellcolor[RGB]{255, 248, 174}0.0245  &  \cellcolor[RGB]{255, 248, 174}0.5701  \\
   3DGS+USD \tiny{(iNGP+NeuS)}
   &  \cellcolor[RGB]{255, 153, 153}0.0175  &  \cellcolor[RGB]{255, 153, 153}0.6342  \\
   \bottomrule
\end{tabular}
}
\vspace{-4pt}
\end{table}

\begin{table}[h]
\centering
\setlength{\tabcolsep}{3pt}
\caption{The quantitative comparison in novel view synthesis. We report PSNR, SSIM, LPIPS on the GSO dataset.}
\label{tab:nvs}
\resizebox{0.99\linewidth}{!}{
\begin{tabular}{lccc}
   \toprule
   Method  & PSNR$\uparrow$ & SSIM$\uparrow$ & LPIPS$\downarrow$  \\
   \midrule
   Realfusion~\cite{melas2023realfusion}    
   & 15.26 & 0.722 & 0.283   \\
   Zero-1-to-3~\cite{liu2023zero}    
   & 18.93 & 0.779 & 0.166   \\
   SyncDreamer~\cite{liu2023syncdreamer}   
   & 20.05 & 0.798 & 0.146 \\
   Wonder3D~\cite{long2023wonder3d}
   & \cellcolor[RGB]{255, 153, 153}26.07 & \cellcolor[RGB]{255, 204, 153}0.924 & \cellcolor[RGB]{255, 248, 174}0.065 \\
   \hline           
   3DGS+USD (3 min)
   & \cellcolor[RGB]{255, 248, 174}25.20 & \cellcolor[RGB]{255, 248, 174}0.917 & \cellcolor[RGB]{255, 204, 153}0.050 \\
   3DGS+USD (6 min)
   & \cellcolor[RGB]{255, 204, 153}25.84 & \cellcolor[RGB]{255, 153, 153}0.925 & \cellcolor[RGB]{255, 153, 153}0.048 \\
   \bottomrule
\end{tabular}
}
\vspace{-8pt}
\end{table}

\section{Analysis of Unbiased Score Distillation}
We have carefully reviewed the discussion in SJC~\cite{wang2023score}  regarding the challenge of \textbf{Out-of-Distribution (OOD)} when using a pretrained denoiser. In response, we have conducted a more in-depth analysis of the role of $\lambda$ in Formula $\color{red}{6}$ of the main paper and the underlying principles behind it, providing a more detailed comparison between SDS~\cite{poole2022dreamfusion}, DDS~\cite{hertz2023delta}, CSD~\cite{yu2023text} and VSD~\cite{wang2023prolificdreamer}:

\begin{description}
\item[SDS:] $w\textcolor{blue}{\Delta_C}+\textcolor{red}{\Delta_O}-\epsilon$

Due to the OOD issue, the residual $(\Delta_O-\epsilon)$ is generally non-zero, which introduces artifacts in the image. For reducing the impact of $(\Delta_O-\epsilon)$, DreamFusion~\cite{poole2022dreamfusion} sets a large guidance weight $(w = 100)$ to improve the sampling quality.

\item[DDS:] $w(\textcolor{blue}{\Delta_{C_{\textbf{edit}}}}-\textcolor{blue}{\Delta_{C_{\textbf{orig}}}})$

The residual component that causes over-smoothing and over-saturation in SDS is removed, allowing DDS to generate high-quality results.

\item[VSD:] $w\textcolor{blue}{\Delta_C}+\textcolor{red}{\Delta_O}-\textcolor{red}{\Delta_O(\text{LoRA})}$

The rendered images during the optimization are OOD for the original pretrained model distribution, but are in-domain for $\epsilon_{\text{LoRA}}$ which simply predicts $\Delta_O(\text{LoRA})$. As a result, mitigating the impact of $\Delta_O$.

\item[USD/CSD:] $w\textcolor{blue}{\Delta_C}+\lambda(\textcolor{red}{\Delta_O}-\epsilon)$

In our approach, we reparameterized SDS and aim to achieve a similar effect by tuning $\lambda$ to mitigate the OOD.

\end{description}

\noindent where $\textcolor{blue}{\Delta_C}=\epsilon_\phi(\mathbf{z_{t}}; y, t)-\epsilon_\phi(\mathbf{z_{t}}; \varnothing, t)$ can be regarded as an optimization direction toward the condition $y$; $\textcolor{red}{\Delta_O}=\epsilon_\phi(\mathbf{z_{t}}; \varnothing, t)$ since directly evaluating the denoiser $\epsilon_\phi$ on rendered images $\mathbf{z_{t}}$ to compute the score leads to OOD and the residual component $(\textcolor{red}{\Delta_O}-\epsilon)$ that leads to over-smoothed and over-saturated.

\section{More Details in Main Paper}

\noindent{\bf Different Viewing Angle Comparisons.}
Our approach demonstrates significant strengths: \textbf{(1)}. viewing angle flexibility: unlike Wonder3D, which is limited to near-horizontal viewing angles and underperforms outside this range, our method supports a much wider range of viewing directions (Fig.~\ref{view_compare}); \textbf{(2)}. target view generation: our method generates reasonable target views for any given relative pose, in contrast to Wonder3D's restriction to six fixed relative poses. These advantages, crucial for diverse applications, highlight our method's unique contributions beyond what is reflected in the current metrics.

\begin{figure}[h]
    \centering
    \includegraphics[width=1.0\linewidth]{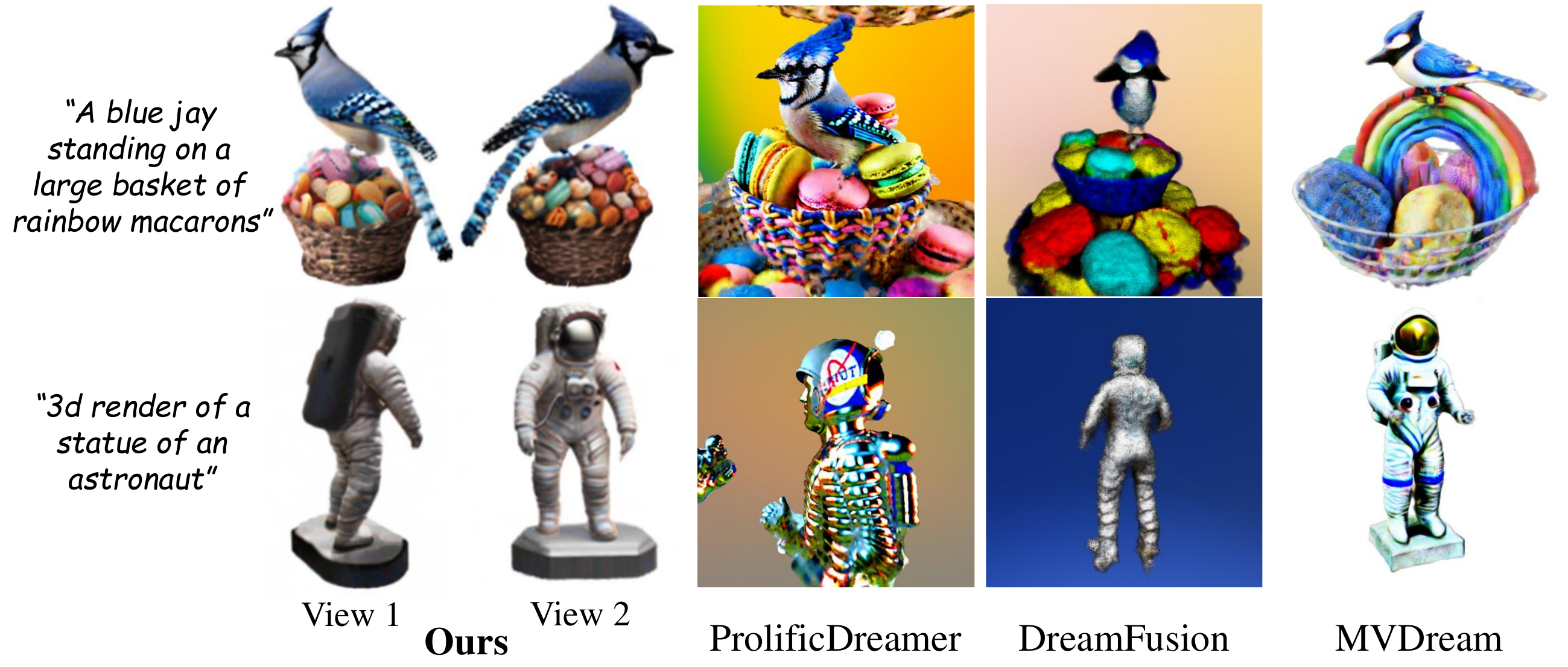}
    \caption{Qualitative results of various text to 3D approaches.}
    \label{fig:Textto3D}
\end{figure}

\noindent{\bf Text-to-Image-to-3D.}
As a case study, we demonstrate that our Unbiased Score Distillation (USD) and view and geometry refinement strategy effectively bridges the gap between multi-view diffuser with text prompts. We use the Stable Diffusion model to generate an image from a text prompt and use our scheme to generate geometry from this image. The result shows a significant advantage of our method in mitigating the multi-face Janus problem and can generate more realistic results, as shown in Fig.~\ref{fig:Textto3D}.

\begin{figure*}[t]
    \centering
    \includegraphics[width=1.0\linewidth]{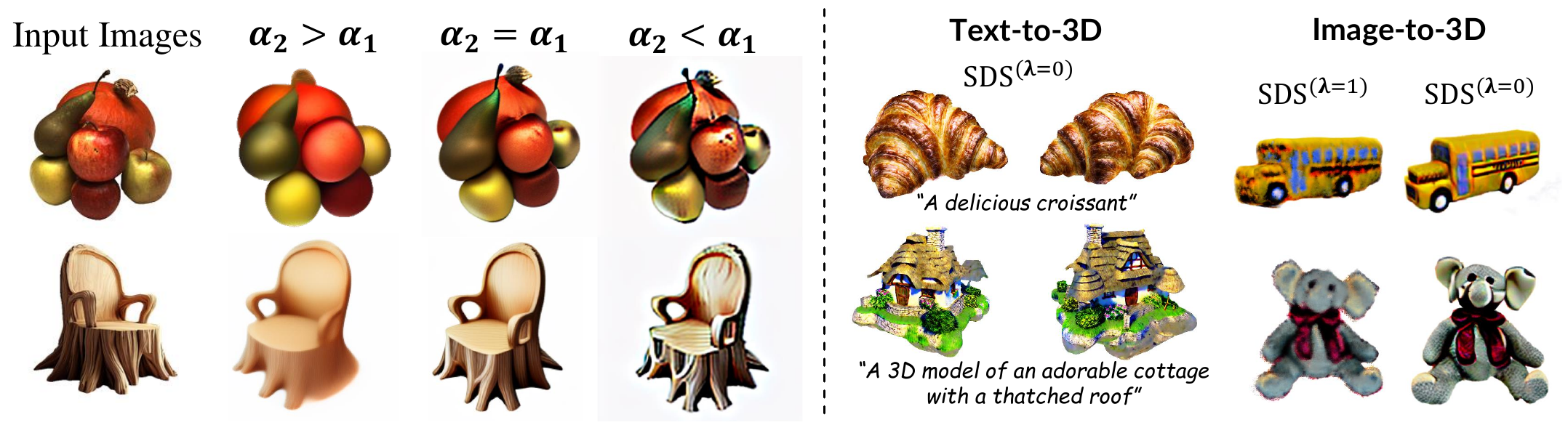}
    \caption{(Left) We compare the effect of the relationship between $\alpha_1$ and $\alpha_2$ on SDS, which can achieve the best when $\alpha_1=\alpha_2$. (Right) Experiments on both text-to-3D and image-to-3D tasks demonstrate that SDS can achieve more detailed results when $\boldsymbol{\lambda}=0$.
    }
    \label{alpha_compare}
\end{figure*}

\begin{figure*}[t]
    \centering
    \includegraphics[width=1.0\linewidth]{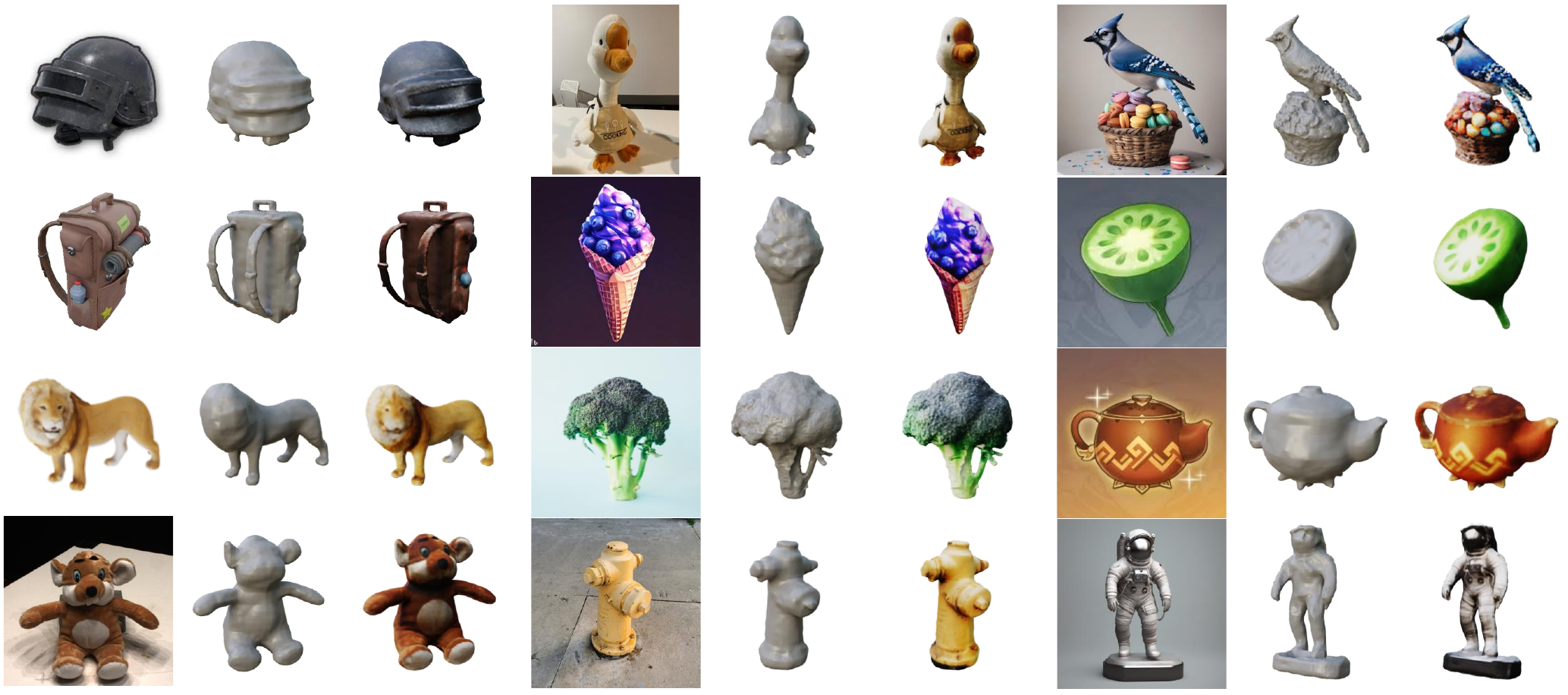}
    \caption{More results. Our model is generally applicable to various types of 2D images.}
    \label{fig:more}
\end{figure*}

\begin{figure}[t]
    \centering
    \includegraphics[width=1.0\linewidth]{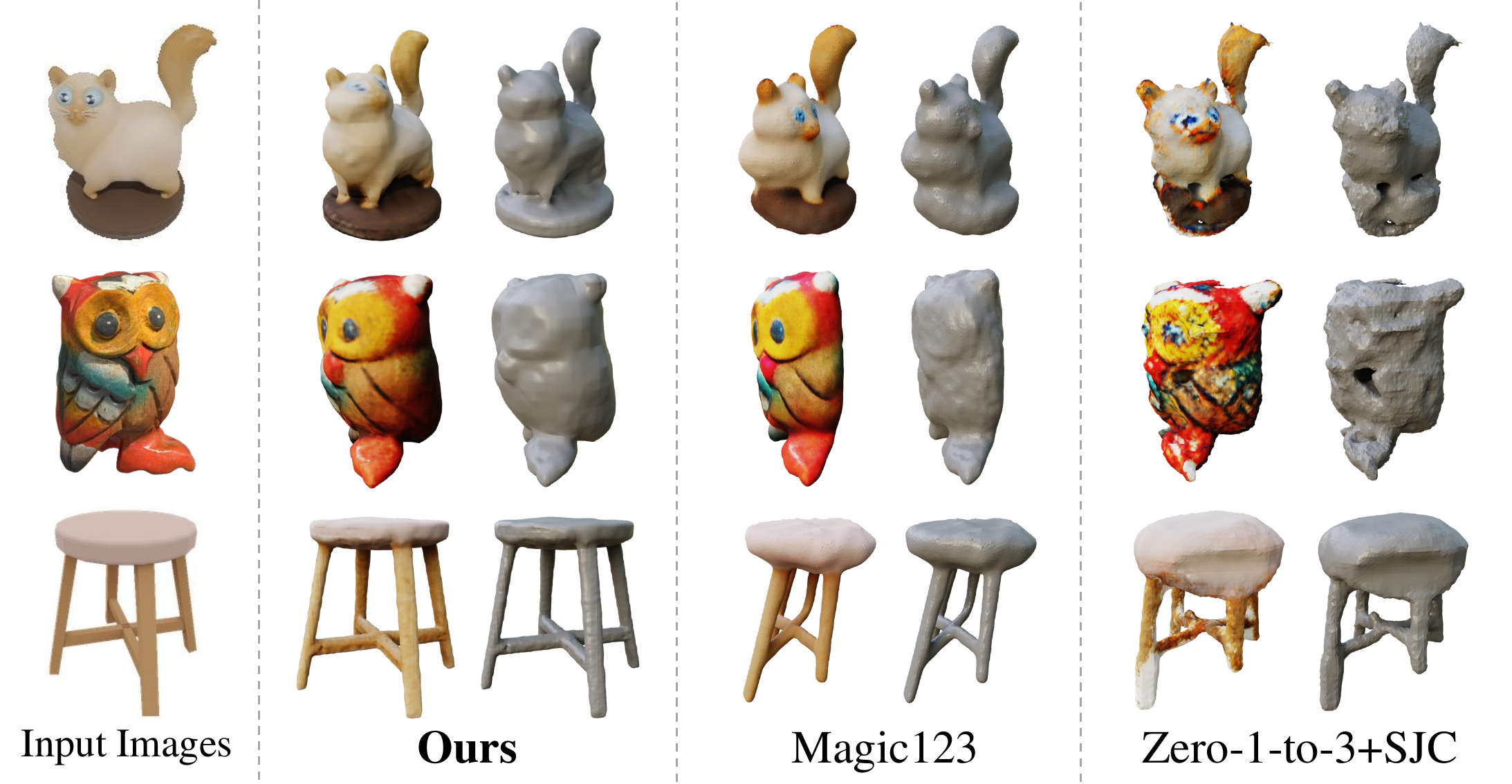}
    \caption{The qualitative comparisons with Magic123 and Zero-1-to-3 (using SJC) in terms of the reconstructed textured meshes.}
    \label{Zero123_Magic123}
\end{figure}

\noindent{\bf SDS/SJC Method.}
We compare our geometry and texture distillation method based USD with Magic123~\cite{qian2023magic123} and Zero-1-to-3~\cite{liu2023zero} (using SJC). Though the results of Magic123 tend to be smooth and clear, it heavily relies on the estimated depth from the input view, which may lead to incorrect estimations when the depth estimator fails. Zero-1-to-3 uses SJC for geometry and texture reconstruction, its formation is biased and hence lead to bad estimations. Ours, however, generate consistent and faithful estimations, as shown in Fig.~\ref{Zero123_Magic123}.

\noindent{\bf The Setting of $\alpha_1$ and $\alpha_2$.}
We provide an empirical validation as in Fig.~\ref{alpha_compare} (Left).

\noindent{\bf The Variable $\boldsymbol{\lambda}$.}
We found that setting $\boldsymbol{\lambda}=0$ can significantly improve the details of the 3D details generated using SDS, as shown in Fig.~\ref{alpha_compare} (Right).

\section{More Results}

We present more reconstruction results in Fig.~\ref{fig:more}. The readers may refer to supplementary videos for $360^\circ$ visualization.

\end{document}